%% file: ncRBMtrack.tex
\documentclass[12pt]{article}
\usepackage{amsmath}
\usepackage{amsfonts}
\usepackage{amssymb}
\usepackage{times}
\usepackage{color}
\usepackage{multirow}

\usepackage[pdftex]{graphicx} 
\usepackage{subfigure} 
\usepackage{algorithm}
\usepackage[noend]{algorithmic}

\usepackage{graphicx}
\usepackage{tabularx}

\usepackage[authoryear]{natbib}
\usepackage{rotating}
\usepackage{bbm}
\usepackage{latexsym}

\newcommand{\captionfonts}{\normalsize}

\makeatletter  
\long\def\@makecaption#1#2{%
  \vskip\abovecaptionskip
  \sbox\@tempboxa{{\captionfonts #1: #2}}%
  \ifdim \wd\@tempboxa >\hsize
    {\captionfonts #1: #2\par}
  \else
    \hbox to\hsize{\hfil\box\@tempboxa\hfil}%
  \fi
  \vskip\belowcaptionskip}
\makeatother   

\def\capstyle#1{\small \emph{#1}}
\newcommand{\xv}{\mathbf{x}}

\newcommand{\hv}{\mathbf{h}}
\newcommand{\cv}{\mathbf{c}}
\newcommand{\av}{\mathbf{a}}
\newcommand{\bv}{\mathbf{b}}
\newcommand{\dv}{\mathbf{d}}
\newcommand{\vv}{\mathbf{v}}
\newcommand{\zv}{\mathbf{z}}
\newcommand{\Vv}{\mathbf{V}}
\newcommand{\Wv}{\mathbf{W}}
\newcommand{\Pv}{\mathbf{P}}

\renewcommand{\vec}[1]{\boldsymbol{\mathrm{#1}}}
\newcommand{\T}{\mathrm{T}}
\renewcommand{\exp}[1]{\operatorname{\mathrm{exp}}\left(#1\right)}
\renewcommand{\algorithmicrequire}{\textbf{Input:}}
\renewcommand{\algorithmiccomment}[1]{\hfill// #1}

\newcounter{algorithmsectioncounter}
\renewcommand{\algorithmicloop}{}
\newcommand{\ALGORITHMSECTION}[1]{\stepcounter{algorithmsectioncounter}\STATE
  \textbf{\arabic{algorithmsectioncounter}.  #1} \LOOP}
\newcommand{\ENDALGORITHMSECTION}{\ENDLOOP}

\begin{document}
\hspace{13.9cm}1

\ \vspace{20mm}\\

\begin{center}
  {\LARGE Learning where to Attend with Deep Architectures for Image Tracking}
\end{center}

\ \\
\textbf{%
\large{Misha Denil}$^{1}$,
\large{Loris Bazzani}$^{2}$,
\large{Hugo Larochelle}$^{3}$}
and
\textbf{\large{Nando de Freitas}$^{1}$}\\
{$^{1}$University of British Columbia.}\\
{$^{2}$University of Verona.}\\
{$^{3}$University of Sherbrooke.}\\

%

\textbf{Keywords:} Restricted Boltzmann machines, Bayesian
optimization, bandits, attention, deep learning, particle filtering,
saliency

\thispagestyle{empty}
\markboth{}{Learning where to Attend}
\ \vspace{-0mm}\\
%
\begin{center} {\bf Abstract} \end{center} We discuss an attentional
model for simultaneous object tracking and recognition that is driven
by gaze data. Motivated by theories of perception, the model consists
of two interacting pathways: identity and control, intended to mirror
the what and where pathways in neuroscience models. The identity
pathway models object appearance and performs classification using
deep (factored)-Restricted Boltzmann Machines. At each point in time
the observations consist of foveated images, with decaying resolution
toward the periphery of the gaze. The control pathway models the
location, orientation, scale and speed of the attended object. The
posterior distribution of these states is estimated with particle
filtering. Deeper in the control pathway, we encounter an attentional
mechanism that learns to select gazes so as to minimize tracking
uncertainty. Unlike in our previous work, we introduce gaze selection
strategies which operate in the presence of partial information and on
a continuous action space.  We show that a straightforward extension
of the existing approach to the partial information setting results in
poor performance, and we propose an alternative method based on
modeling the reward surface as a Gaussian Process.  This approach
gives good performance in the presence of partial information and
allows us to expand the action space from a small, discrete set of
fixation points to a continuous domain.

\input{introduction}
\input{identity-pathway}
\input{control-pathway}
\input{gaze-control}
\input{algorithm}
\input{experiments}
\input{conclusion}

\section*{Acknowledgments} 
We thank Ben Marlin, Kenji Okuma, Marc'Aurelio Ranzato and Kevin
Swersky. This work was supported by CIFAR's NCAP program and NSERC.

{
\bibliographystyle{icml2011}
\bibliography{kevinbib,nips,biblio}
}

\end{document}

%% file: introduction.tex
\section{Introduction}
\label{sec:intro}

Humans track and recognize objects effortlessly and efficiently,
exploiting attentional mechanisms~\citep{Rensink:2000,Colombo:2001} to
cope with the vast stream of data.  We use the human visual system as
inspiration to build a system for simultaneous object tracking and
recognition from gaze data.  An attentional strategy is learned online
to choose fixation points which lead to low uncertainty in the
location of the target object.  Our tracking system is composed of two
interacting pathways.  This separation of responsibility is a common
feature in models from the computational neuroscience literature as it
is believed to reflect a separation of information processing into
ventral and dorsal pathways in the human
brain~\citep{olshausen1993neurobiological}.

\begin{figure}[h!]
 \begin{center}
   \includegraphics[width=1\textwidth]{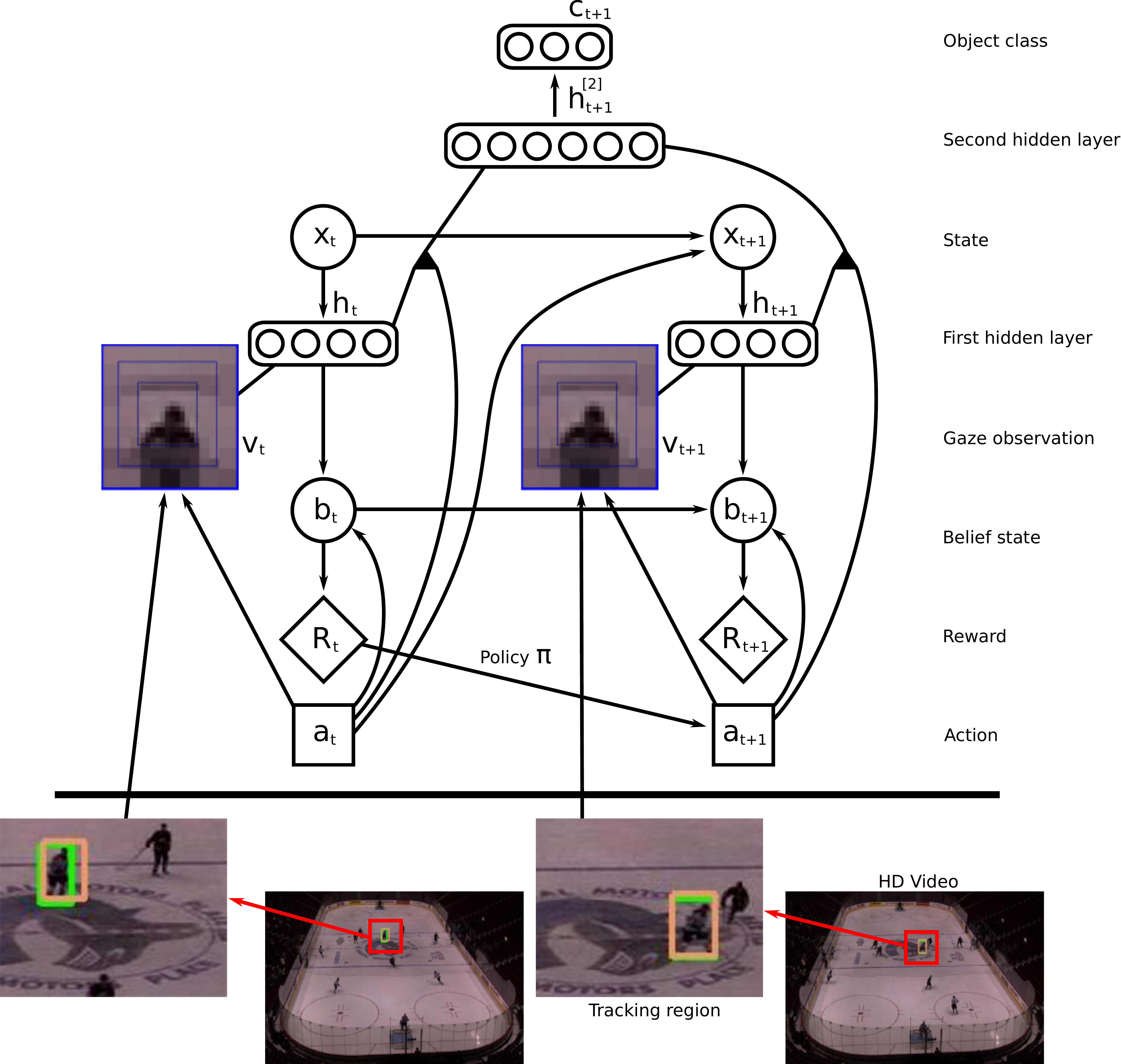}
 \end{center}
 \vspace{-5pt}
 \caption{\capstyle{From a sequence of gazes
     $(\mathbf{v}_t,\mathbf{v}_{t+1},\ldots)$, the model infers the
     hidden features $\mathbf{h}$ for each gaze (that is, the
     activation intensity of each receptive field), the hidden
     features for the fusion of the sequence of gazes and the object
     class $\mathbf{c}$. Only one time step of classification is kept
     in the figure for clarity. The location, size, speed and
     orientation of the gaze patch are encoded in the state
     $\mathbf{x}_t$. The actions $\mathbf{a}_t$ follow a learned
     policy $\mathbf{\pi}_t$ that depends on the past rewards $\{r_1,
     \ldots, r_{t-1}\}$. This particular reward is a function of the
     belief state $\mathbf{b}_t =
     p(\mathbf{x}_t|\mathbf{a}_{1:t},\mathbf{h}_{1:t})$, also known as
     the filtering distribution. Unlike typical commonly used
     partially observed Markov decision models (POMDPs), the reward is
     a function of the beliefs. In this sense, the problem is closer
     to one of sequential experimental design.
     With more layers in the ventral $\vv - \hv - \hv^{[2]} - \cv$
     pathway, other rewards and policies could be designed to
     implement higher-level attentional strategies.  }} \vspace{-3mm}
\label{fig:graphmod}
\end{figure}

The \emph{identity} pathway (ventral) is responsible for comparing
observations of the scene to an object template using an appearance
model, and on a higher level, for classifying the target object.  The
identity pathway consists of a two hidden layer deep network.  The top
layer corresponds to a multi-fixation Restricted Boltzmann Machine
(RBM) \citep{Larochelle2010}, as shown in
Figure~\ref{fig:graphmod}. It accumulates information from the first
hidden layers at consecutive time steps. For the first layers, we use
(factored)-RBMs~\citep{hinton2006rdd,Ranzato2010,welling2005efh,swersky2011autoencoders},
but autoencoders \citep{vincent2008eac}, sparse
coding~\citep{Olshausen1996, Kavukcuoglu2009}, two-layer
ICA~\citep{Koster2007} and convolutional
architectures~\citep{lee2009convolutional} could also be adopted.

\begin{figure}[t!]
 \begin{center}
   \includegraphics[width=0.25\textwidth]{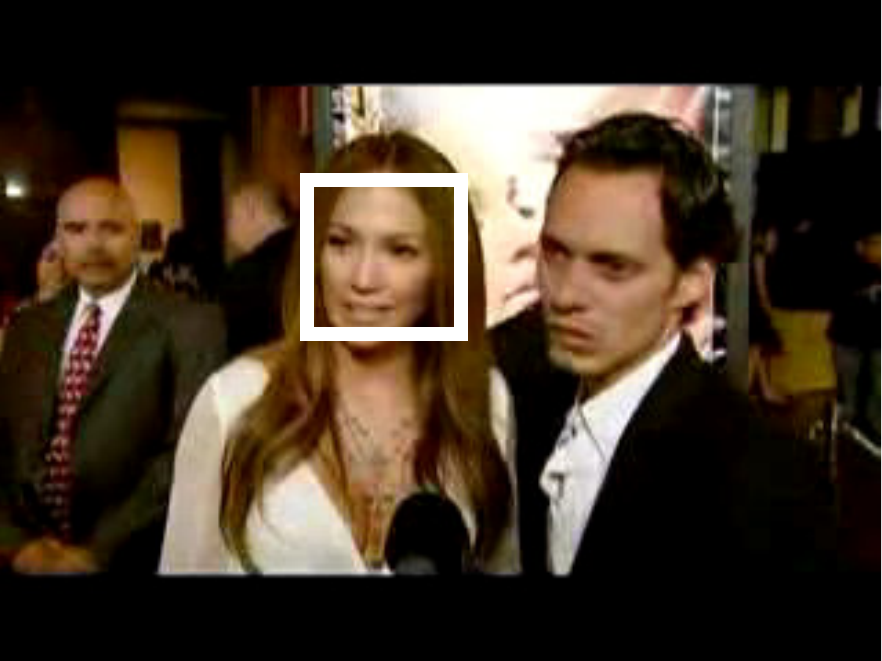}
   \quad
   \includegraphics[width=0.2\textwidth]{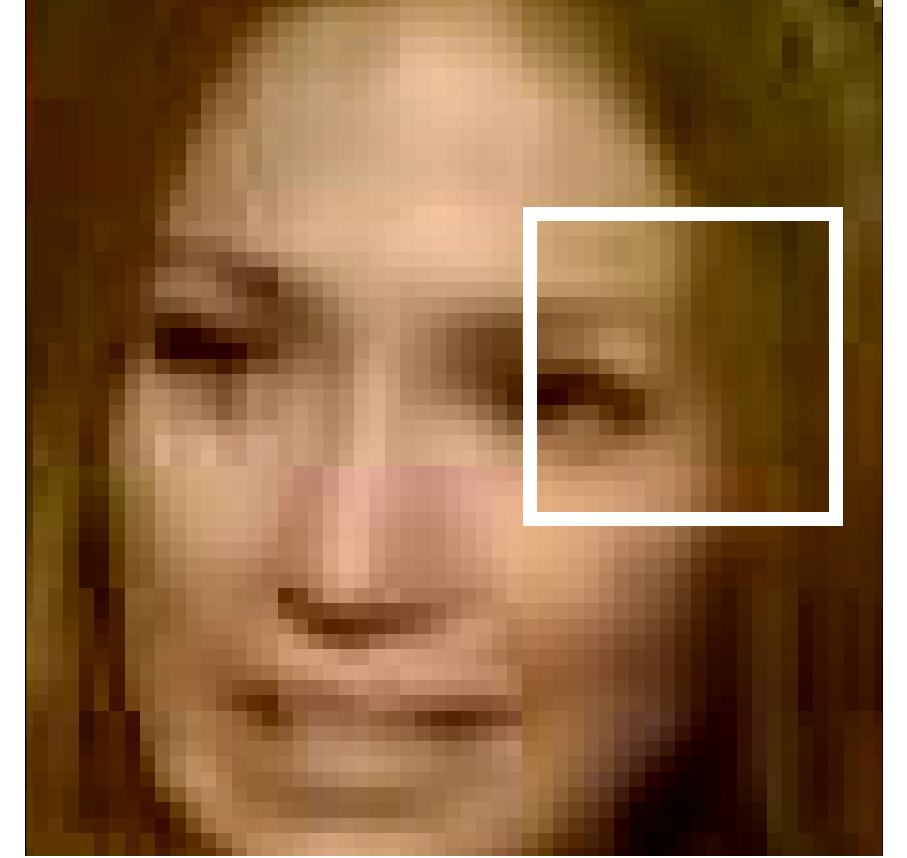}
   \quad
   \includegraphics[width=0.2\textwidth]{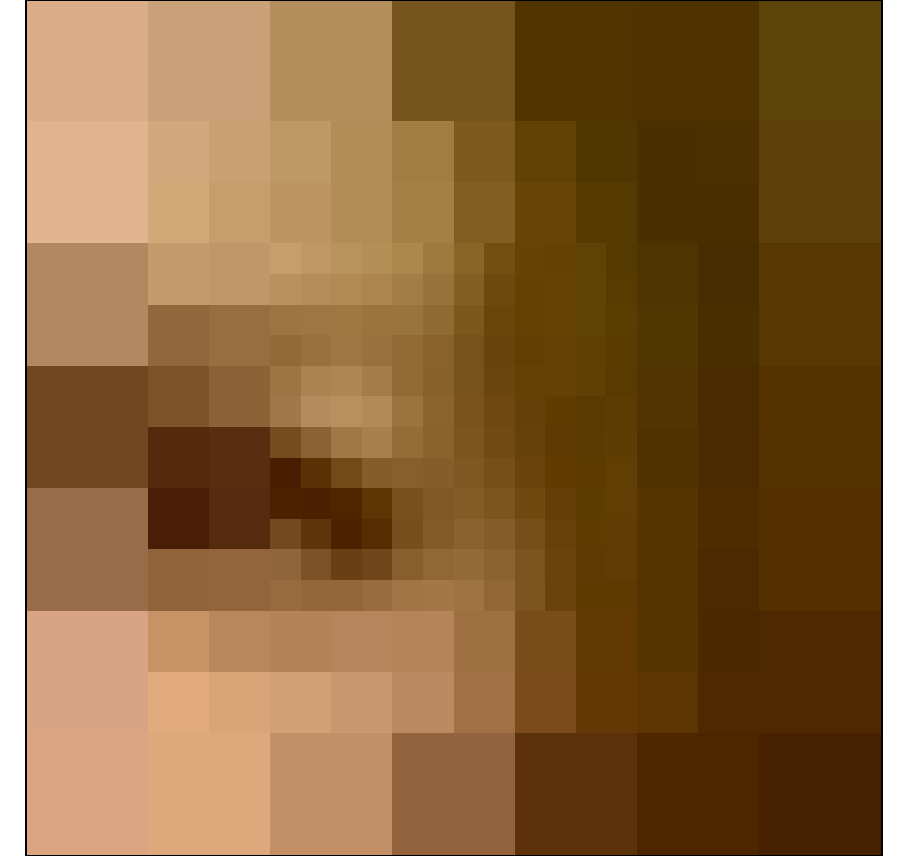}
   \quad
   \includegraphics[width=0.2\textwidth]{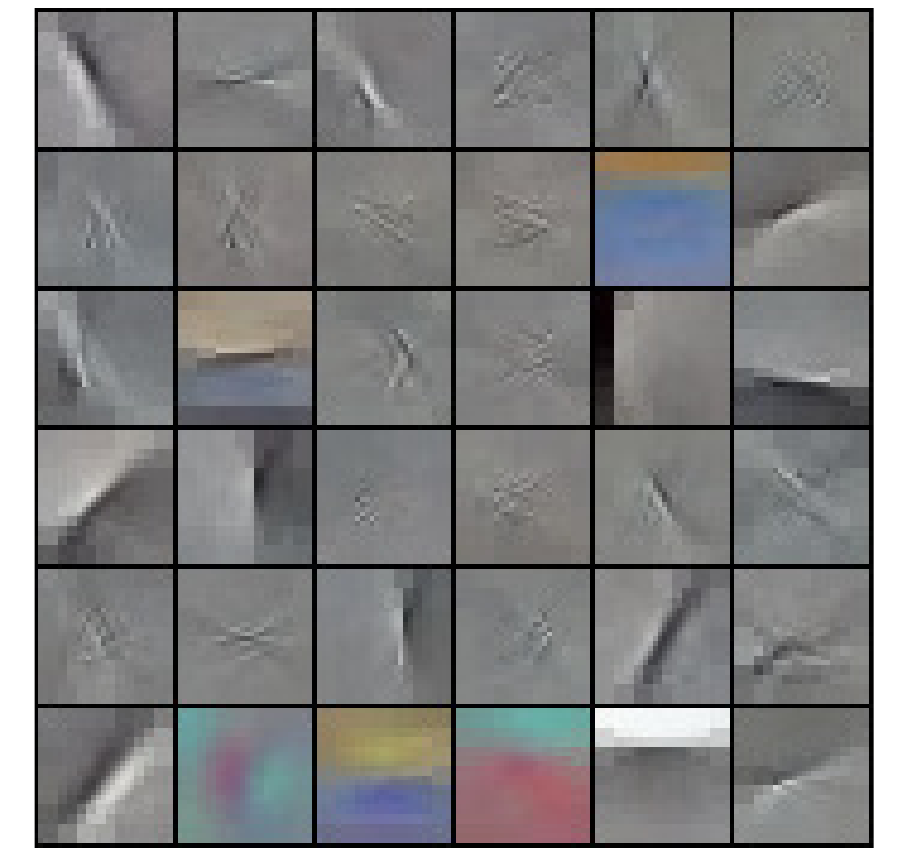}
 \end{center}
 \vspace{-10pt}
 \caption{\capstyle{\textbf{Left:} A typical video frame with the
     estimated target region highlighted.  To cope with the large
     image size our system considers only the target region at each
     time step. \textbf{Centre left:} A close-up of the template
     extracted from the first frame.  The template is compared to the
     target region by selecting a fixation point for comparison as
     shown.  \textbf{Centre right:} A visualization of a single
     fixation.  In addition to covering only a very small portion of
     the original frame, the image is foveated with high resolution
     near the centre and low resolution on the periphery to further
     reduce the dimensionality.  \textbf{Right:} The most active
     filters of the first layer (factored)-RBM when observing the
     displayed location.  The control pathway compares these features
     to the features active at the corresponding scene location in
     order to update the belief state.}}
 \label{fig:template-example}
\end{figure}

The \emph{control} pathway (dorsal) is responsible for aligning the
object template with the full scene so the remaining modules can
operate independently of the object's position and scale.  This
pathway is separated into a localization module and a fixation module
which work cooperatively to accomplish this goal.  The localization
module is implemented with a particle filter~\citep{315} which
estimates the location, velocity and scale of the target object.  We
make no attempt to implement such states with neural architectures,
but it seems clear that they could be encoded with grid
cells~\citep{mcnaughton2006} and retinotopic maps as in V1 and the
superior colliculus~\citep{Rosa2002,Girard2005}.  The fixation module
learns an attentional strategy to select fixation points relative to
the object template.  These fixation points are the centres of partial
template observations, and are compared with observations of the
corresponding locations in the scene using the appearance model (see
Figure~\ref{fig:template-example}).  Reward is assigned to each
fixation based on the uncertainty of the target location at each time
step.  The fixation module uses the reward signal to adapt its gaze
selection policy to achieve good localization.  Our previous
work~\citep{bazzani2010learning} used Hedge~\citep{auer1998gambling,
  Freund:1997} to learn this policy.  In this extended paper we show
that a straightforward adaptation of our previous approach to the
partial information setting results in poor performance, and we
propose an alternative method based on modelling the reward surface as
a Gaussian Process.  This approach gives good performance in the
presence of partial information and allows us to expand the action
space from a small, discrete set of fixation points to a continuous
domain.

The proposed system can be motivated from different
perspectives. First, starting with~\cite{93}, many particle filters
have been proposed for image tracking, but these typically use simple
observation models such as B-splines~\citep{93} and colour
templates~\citep{Okuma:2004}. RBMs are more expressive models of
shape, and hence, we conjecture that they will play a useful role
where simple appearance models fail. Second, from a deep learning
computational perspective, this work allows us to tackle large images
and video, which is typically not possible due to the number of
parameters required to represent large images in deep models. The use
of fixations synchronized with information about the state
(e.g. location and scale) of such fixations eliminates the need to
look at the entire image or video. Third, the system is invariant to
image transformations encoded in the state, such as location, scale
and orientation. Fourth, from a dynamic sensor network perspective,
this paper presents a very simple, but efficient and novel way of
deciding how to gather measurements dynamically.  Lastly, in the
context of psychology, the proposed model realizes to some extent the
functional architecture for dynamic scene representation
of~\cite{Rensink:2000}. The rate at which different attentional
mechanisms develop in newborns (including alertness, saccades and
smooth pursuit, attention to object features and high-level task
driven attention) guided the design of the proposed approach and was a
great source of inspiration~\citep{Colombo:2001}.

Our attentional model can be seen as building a saliency
map~\citep{koch1985shifts} over the target template.  Previous work on
saliency modelling has focused on identifying salient points in an
image using a bottom up process which looks for outliers under some
local feature model (which may include a task dependent prior, global
scene features, or various other heuristics).  These features can be
computed from static images~\citep{torralba2006contextual}, or from
local regions of spacetime~\citep{gaborski2004detection} for video.
Additionally, a wide variety of different feature types have been
applied to this problem, including engineered
features~\citep{gao2007discriminant} as well as features that are
learned from data~\citep{zhang2009sunday}.  Core to these methods is
the idea that saliency is determined by some type of novelty measure.
Our approach is different, in that rather than identifying locally or
globally novel features, our process identifies features which are
useful for the task at hand.  In our system the saliency signal for a
location comes from a top down process which evaluates how well the
features at that location enable the system to localize the target
object.  The work of~\cite{gao2007discriminant} considers a similar
approach to saliency by defining saliency to be the mutual information
between the features at a location and the class label of an object
being sought; however, in order to make their model tractable the
authors are forced to use specifically engineered features.  Our
system is able to cope with arbitrary feature types, and although we
consider only on localization in this paper, our model is sufficiently
general to be applied to identifying salient features for other goals.

Recently, a dynamic RBM state-space model was proposed
in~\cite{Taylor2010}. Both the implementation and intention behind
that proposal are different from the approach discussed here. To the
best of our knowledge, our approach is the first successful attempt to
combine dynamic state estimation from gazes with online policy
learning for gaze adaptation, using deep network network models of
appearance. Many other dual-pathway architectures have been proposed
in computational neuroscience, including~\cite{Olshausen1993}
and~\cite{Postma1997}, but we believe ours has the advantage that it
is very simple, \emph{modular} (with each module easily replaceable),
suitable for large datasets and easy to extend.


%% file: identity-pathway.tex
\section{Identity Pathway}
\label{sec:identity}

The identity pathway in our model mirrors the ventral pathway in
neuroscience models.  It is responsible for modelling the appearance
of the target object and also, at a higher level, for classification.

\subsection{Appearance Model}
\label{sec:learningapp}

\begin{figure}[t!]
 \begin{center}
   \includegraphics[width=0.75\textwidth]{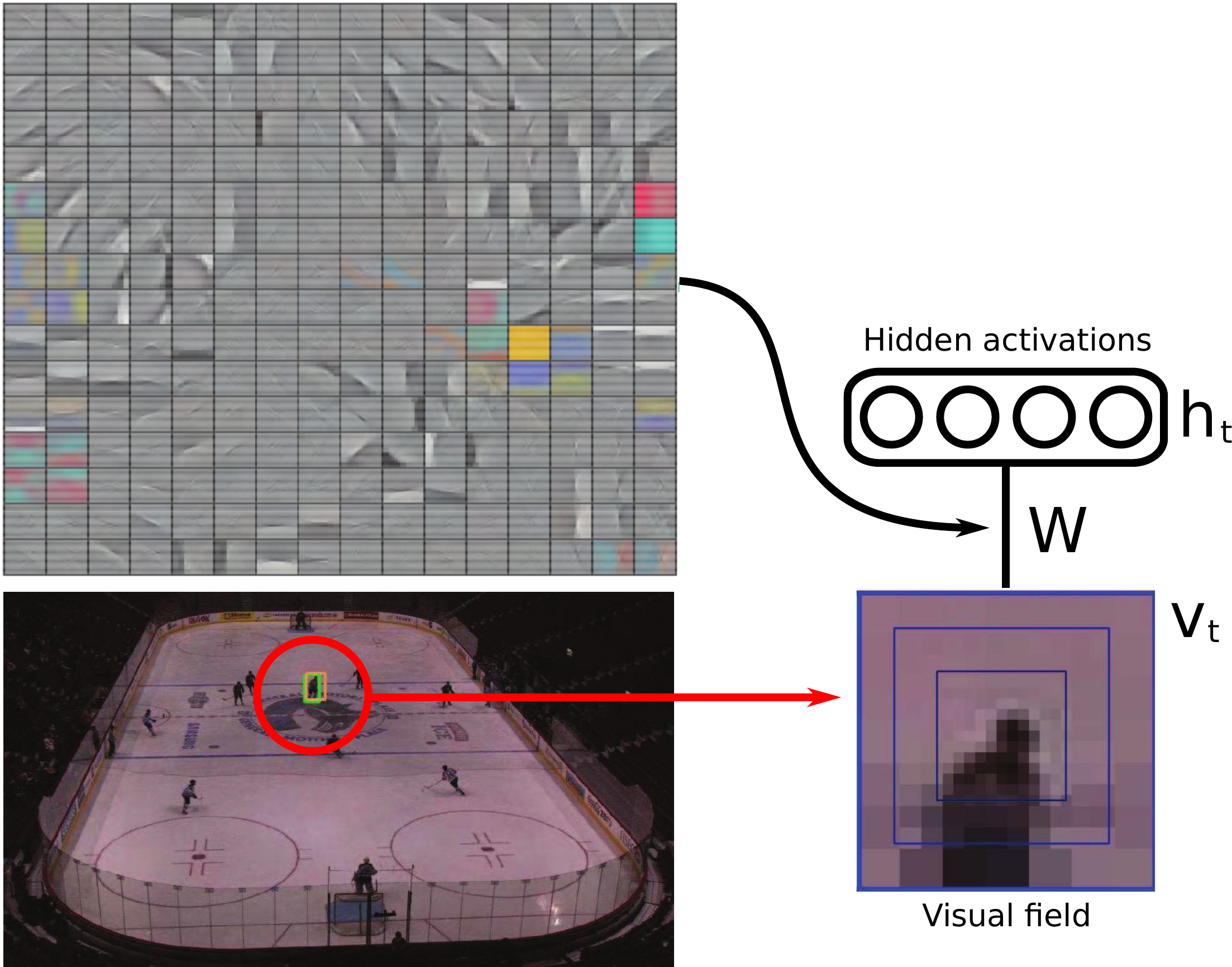}
 \end{center}
 \vspace{-10pt}
 \caption{\capstyle{An RBM senses a small foveated image derived from
     the video.  The level of activation of each filter is recorded in
     the $\vec{h}_t$ units.  The RBM weights (filters) $\vec{W}$ are
     visualized in the upper left.  We currently pre-train these
     weights.}}
 \label{fig:appearance-model}
\end{figure}

We use (factored)-RBMs to model the appearance of objects and perform
object classification using the gazes chosen by the control module
(see Figure~\ref{fig:appearance-model}). These undirected
probabilistic graphical models are governed by a Boltzmann
distribution 
over the gaze data $\vv_t$ and the hidden features $\hv_t \in
\{0,1\}^{n_h}$. We assume that the receptive fields $\mathbf{w}$, also
known as RBM weights or filters, have been trained beforehand. We also
assume that readers are familiar with these models and, if otherwise,
refer them to~\cite{Ranzato2010} and~\cite{Swersky2010}.

\subsection{Classification Model}

\begin{figure}[t!]
 \begin{center}
   \includegraphics[width=0.75\textwidth]{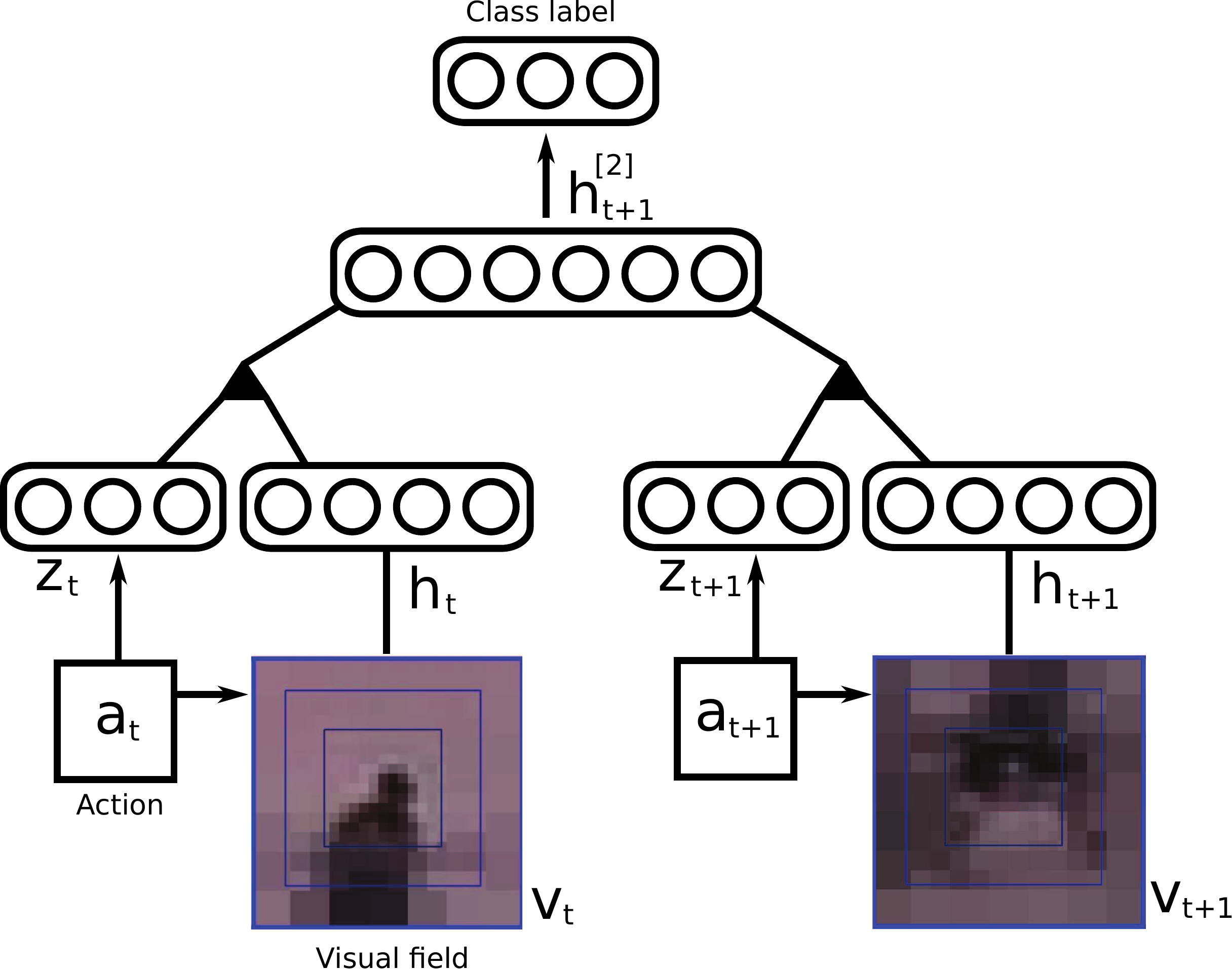}
 \end{center}
 \vspace{-10pt}
 \caption{\capstyle{Gaze accumulation and classification in the
     identity pathway. A multi-fixation RBM models the conditional
     distribution (given the gaze positions $\av_t$) of $\Delta$
     consecutive hidden features $\hv_t$, extracted by the first layer
     RBM on the foveated images. In this illustration, $\Delta=2$. The
     multi-fixation RBM encodes the gaze position $\av_t$ in a ``one
     hot'' representation noted $\zv_t$. The activation probabilities
     of the second layer hidden units $\hv_t^{[2]}$ are used by a
     classifier to predict the object's class.}}
 \label{fig:gaze-accumulation}
\end{figure}

The identity pathway also performs object recognition, classifying a
sequence of gaze instances selected with the gaze policy. We implement
a multi-fixation RBM very similar to the one proposed in
\cite{Larochelle2010}, where the binary variables $\mathbf{z}_t$ (see
Figure~\ref{fig:gaze-accumulation}) are introduced to encode the
relative gaze location $\mathbf{a}_t$ within the multi-fixation RBM (a
``1 in $K$'' or ``one hot'' encoding of the gaze location was used for
$\mathbf{z}_t$).

The multi-fixation RBM uses the relative gaze location
information in order to aggregate the first hidden layer
representations $\hv_t$ at $\Delta$ consecutive time steps into a
single, higher level representation $\hv^{[2]}_t$.

More specifically, the energy function of the multi-fixation RBM is:
\begin{align*}
  E(\hv_{t-\Delta+1:t},\zv_{t-\Delta+1:t},\hv^{[2]}_t) = -\dv^\top
  \hv^{[2]}_t -\sum_{i=1}^{\Delta} \bv^\top \hv_{t-\Delta+i} +
  \sum_{f=1}^{F} (\Pv_{f,:} \hv^{[2]}_t) (\Wv_{f,:} \hv_{t-\Delta+i})
  (\Vv_{f,:} \zv_{t-\Delta+i})
\end{align*}
where the notation $\Pv_{f,:}$ refers to the $f^{\rm th}$ row vector
of the matrix $\vec{P}$.  From this energy function, we define a
distribution over $\hv_{t-\Delta+1:t}$ and $\hv^{[2]}_t$ (conditioned
on $\zv_{t-\Delta+1:t}$) through the Boltzmann distribution:
\begin{equation}
  p(\hv_{t-\Delta+1:t},\hv^{[2]}_t | \zv_{t-\Delta+1:t}) = \exp{-E(\hv_{t-\Delta+1:t},\zv_{t-\Delta+1:t},\hv^{[2]}_t)} / Z( \zv_{t-\Delta+1:t}) \label{eqn:mfrbm}
\end{equation}
where the normalization constant $Z(\zv_{t-\Delta+1:t})$ ensures that
Equation~\ref{eqn:mfrbm} sums to 1.  To sample from this distribution,
one can use Gibbs sampling by alternating between sampling the
top-most hidden layer $\hv^{[2]}_t$ given all individual processed
gazes $\hv_{t-\Delta+1:t}$ and vice versa. To train the multi-fixation
RBM, we collect a training set consisting in sequences of $\Delta$
pairs $(\hv_t,\zv_t)$ by randomly selecting $\Delta$ gaze positions at
which to fixate and computing the associated $\hv_t$. These sets are
extracted from a collection of images in which the object to detect
has been centred. Unsupervised learning using
contrastive divergence can then be performed on this training set. See
\citet{Larochelle2010} for more details.

The main difference between this multi-fixation RBM and the one
described in \citet{Larochelle2010} is that $\hv^{[2]}_t$ does not
explicitly model the class label $\cv_t$. Instead, a multinomial
logistic regression classifier is trained separately, to predict
$\cv_t$ from the aggregated representation extracted from
$\hv^{[2]}_t$. More specifically, we use the vector of activation
probabilities of all hidden units $h^{[2]}_{t,j}$ in $\hv^{[2]}_t$,
conditioned on $\hv_{t-\Delta+1:t}$ and $\zv_{t-\Delta+1:t}$, as the
aggregated representation:
\begin{align*}
  p(h^{[2]}_{t,j} = 1 | \hv_{t-\Delta+1:t},\zv_{t-\Delta+1:t}) = {\rm
    sigm}\left(d_j + \sum_{i=1}^{\Delta} \sum_{f=1}^{F} \Pv_{f,j}
    (\Wv_{f,:} \hv_{t-\Delta+i}) (\Vv_{f,:} \zv_{t-\Delta+i})\right)
\end{align*}
We experimented with a single fixation module, but found the
multi-fixation module to increase classification accuracy.  To improve
the estimate the class variable $\mathbf{c}_t$ over time, we
accumulate the classification decisions at each time step.

Note that the process of pursuit (tracking) is essential to
classification. As the target is tracked, the algorithm fixates at
locations near the target's estimated location. The size and
orientation of these fixations also depends on the corresponding state
estimates. Note that we don't fixate exactly at the target location
estimate as this would provide only one distinct fixation over several
time steps if the tracking policy has converged to a specific gaze. It
should also be pointed out that instead of using random fixations, one
could again use the control strategy proposed in this paper to decide
where to look with respect to the track estimate so as to reduce
classification uncertainty. We leave the implementation of this extra
attentional mechanism for future work.


%% file: control-pathway.tex
\section{Control Pathway}
\label{sec:control}

\begin{figure}[t!]
 \begin{center}
   \includegraphics[width=0.5\textwidth]{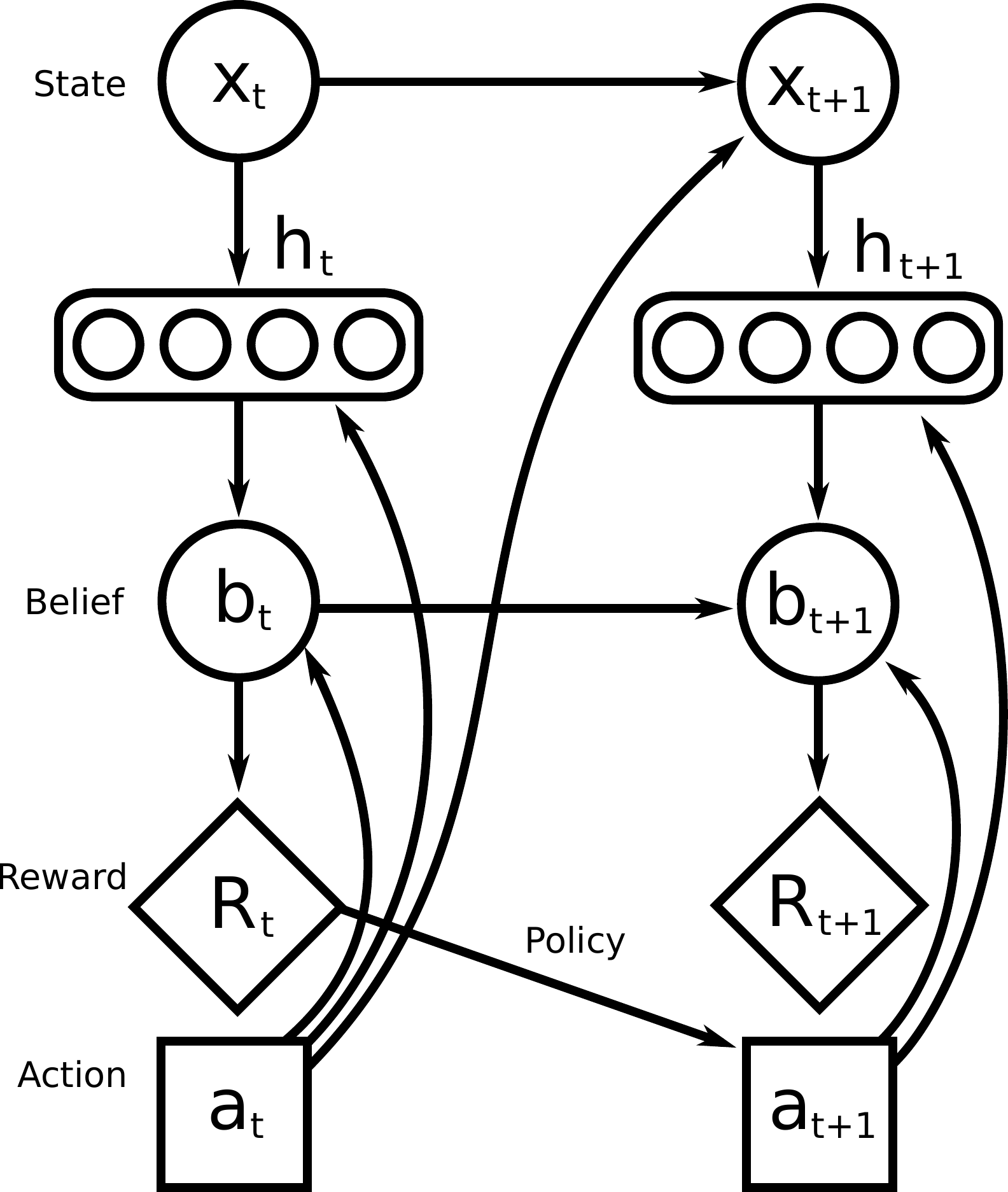}
 \end{center}
 \vspace{-10pt}
 \caption{\capstyle{Influence diagram for the control pathway.  The
     true state of the tacked object $\vec{x}_t$, generates some set
     of features $\vec{h}_t$, in the identity pathway.  These features
     depend on the action chosen at time $t$ and are used to update
     the belief state $\vec{b}_t$.  Statistics of the belief state are
     collected to compute the reward $r_t$, which is used to update
     the policy for the next time step.}}
 \label{fig:control-pathway}
\end{figure}

The control pathway mirrors the responsibility of the dorsal pathway
in human visual processing.  It tracks the state of the target
(position, speed, etc) and normalizes the input so that other modules
need not account for these variations.  At a higher level it is
responsible for learning an attentional strategy which maximizes the
amount of information learned with each fixation.  The structure of
the control pathway is shown in Figure~\ref{fig:control-pathway}.

\subsection{State-space model}
\label{sec:state-space-model}

The standard approach to image tracking is based on the formulation of
Markovian, nonlinear, non-Gaussian state-space models, which are
solved with approximate Bayesian filtering techniques.  In this
setting, the unobserved signal (object's position, velocity, scale,
orientation or discrete set of operations) is denoted
$\left\{\mathbf{x}_{t}\in \mathbf{\cal X};t\in
  \mathbb{N}\right\}$. This signal has initial distribution $p\left(
  \mathbf{x}_{0}\right) $ and transition equation $p\left(
  \left. \mathbf{x}_{t}\right| \mathbf{x}_{t-1},\av_{t-1}\right).$
Here $\av_t \in {\cal A}$ denotes an action at time $t$, defined on a
compact set $\mathcal{A}$.  For descrete policies $\mathcal{A}$ is
finitie whereas for continuous policies $\mathcal{A}$ is a region in
$\mathbb{R}^2$.  The observations $ \left\{ \mathbf{h}_{t} \in
  \mathbf{\cal H};t\in \mathbb{N}^{*}\right\} $, are assumed to be
conditionally independent given the process state $\left\{
  \mathbf{x}_{t};t\in \mathbb{N}\right\} $.  Note that from the state
space model perspective the observations are the hidden units of the
second layer of the of the appearance model in the identity pathway.
In summary, the state-space model is described by the following
distributions:
\begin{eqnarray}
  p\left( \mathbf{x}_{0}\right)  & \nonumber \\
  p\left( \left. \mathbf{x}_{t}\right| \mathbf{x}_{t-1},\av_{t-1}\right) &\text{ for }t\geq 1\nonumber \\
  p\left( \left. \mathbf{h}_{t}\right| \mathbf{x}_{t},\av_t\right) &\text{ for }t\geq 1,\nonumber 
\end{eqnarray}
 For the transition
model, we will adopt a classical autoregressive process.

Our aim is to estimate recursively in time the \emph{posterior
  distribution}\index{Posterior distribution}\footnote{We use the
  notation $\mathbf{x}_{0:t}\triangleq \left\{
    \mathbf{x}_{0},...,\mathbf{x}_{t}\right\}$ to represent the past
  history of a variable over time.} $ p\left(
  \left. \mathbf{x}_{0:t}\right| \mathbf{h}_{1:t},\av_{1:t}\right) $
and its associated features, including the marginal distribution
$\bv_t \triangleq p\left( \left.  \mathbf{x}_{t}\right|
  \mathbf{h}_{1:t},\av_{1:t}\right) $ --- known as the \emph{filtering
  distribution} or \emph{belief state}. This distribution satisfies
the following recurrence:
\[
\bv_t \propto p(\hv_t|\xv_t,\av_t)
\int p(\xv_t|\xv_{t-1},\av_{t-1})
p(d\xv_{t-1}|\hv_{1:t-1},\av_{1:t-1}).
\]
Except for standard distributions (\emph{e.g.}\ Gaussian or discrete),
this recurrence is intractable.

After learning the observation model we will use it for tracking.  The
observation model is often defined in terms of the distance of the
observations from a template $\tau$,
\[
p\left(\mathbf{h}_{t} | \mathbf{x}_{t},\av_t \right) \propto
\exp{-d(\hv({\xv_t,\av_t}),\tau)},
\]
where $d(\cdot,\cdot)$ denotes a distance metric and $\tau$ an object
template (for example, a color histogram or spline).  In this model,
the observation $\hv({\xv_t,\av_t})$ is a function of the current
state hypothesis and the selected action.  The problem with this
approach is eliciting a good template. Often color histograms or
splines are insufficient. For this reason, we will construct the
templates with (factored)-RBMs as follows. First, optical flow is used
to detect new object candidates entering the visual scene.  Second, we
assign a template to the detected object candidate, as shown in
Figure~\ref{fig:template-example}. The same figure also shows a
typical foveated observation (higher resolution in the center and
lower in the periphery of the gaze) and the receptive fields for this
observation learned beforehand with an RBM. The control algorithm will
be used to learn which parts of the template are most informative,
either by picking from amoung a predefined set of fixation points, or
by using a continuous policy.
Finally, we define the likelihood of each observation directly in
terms of the distance of the hidden units of the RBM
$\hv({\xv_t,\av_t,\vv_t})$, to the hidden units of the corresponding
template region $\hv(\xv_1,\av_t=k,\vv_1)$. That is,
\[
p\left(\mathbf{h}_{t} | \mathbf{x}_{t},\av_t=k \right)\propto
\exp{-d(\hv({\xv_t,\av_t=k,\vv_t}), \hv(\xv_1,\av_t=k,\vv_1))}.
\]
The above template is static, but conceivably one could adapt it over
time.

\subsection{Reward Function}

A gaze control strategy specifies a policy $\pi(\cdot)$ for selecting
fixation points.  The purpose of this strategy is to select fixation
points which maximize an instantaneous reward function $r_t(\cdot)$.
The reward can be any desired behaviour for the system, such as
minimizing posterior uncertainty or achieving a more abstract goal. We
focus on gathering observations so as to minimize the uncertainty in
the estimate of the filtering distribution: $ r_t(\vec{a}_t|\bv_t)
\triangleq u[\widetilde{p}(\xv_t|\hv_{1:t},\av_{1:t})]$.  More
specifically, as discussed later, this reward will be a function of
the variance of the importance weights of the particle filter
approximation $\widetilde{p}(\xv_t|\hv_{1:t},\av_{1:t})$ of the belief
state.

It is also useful to consider the cumulative reward
\begin{align*}
  R_T = \sum_{t=1}^T r_t(\vec{a}_t|\vec{b}_t) \enspace,
\end{align*}
which is the sum of the instantaneous rewards which have been received
up to time $T$.  The gaze control strategies we consider are all
``no-regret'' which means that the average gap between our cumulative
reward and the cumulative reward from always picking the optimal
action goes to zero as $T \to \infty$.

In our
current implementation, each action is a different gaze location and
the objective is to choose where to look so as to minimize the
uncertainty about the belief state.


%% file: gaze-control.tex
\section{Gaze control}
\label{sec:gaze-control}

We compare several different strategies for learning the gaze
selection policy.  In an earlier version of this
work~\citep{bazzani2010learning} we learned the gaze selection policy
with a portfolio allocation algorithm called
Hedge~\citep{Freund:1997,Auer:1998}.  Hedge requires knowledge of the
rewards for all actions at each time step, which is not realistic when
gazes must be preformed sequentially, since the target object will
move between fixations.  We compare this strategy, as well as two
baseline methods, to two very different alternatives.

EXP3 is an extension of Hedge to partial information
games~\citep{auer2001nonstochastic}.  Unlike Hedge, EXP3 requires
knowledge of the reward only for the action selected at each time
step.  EXP3 is more appropriate to the setting at hand, and is also
more computationally efficient than Hedge; however, this comes at a
cost of substantially lower theoretical performance.

Both Hedge and EXP3 learn gaze selection policies which choose among a
discrete set of predetermined fixation points.  We can instead learn a
continuous policy by estimating the reward surface using a Gaussian
Process~\citep{rasmussen2006gaussian}.  By assuming that the reward
surface is smooth, we can draw on the tools of Bayesian
optimization~\citep{brochu2010tutorial} to search for the optimal gaze
location using as few exploratory steps as possible.

The following sections describe each of these approaches in more
detail.

\subsection{Baseline}

We consider two baseline strategies, which we call random and
circular.  The random strategy samples gaze selections uniformly from
a small discrete set of possibilities.  The circular strategy also
uses a small discrete set of gaze locations and cycles through them in
a fixed order.

\subsection{Hedge}

To use Hedge~\citep{Freund:1997,Auer:1998} for gaze selection we must
first discretize the action space by selecting a fixed finite number
of possible fixation points.  Hedge maintains an importance weight
$G(i)$ for each possible fixation point and uses them to form a
stochastic policy at each time step.  An action is selected according
to this policy and the reward for each possible action is observed.
These rewards are then used to update the importance weights and the
process repeats.  Pseudo code for Hedge is shown in
Algorithm~\ref{alg:HEDGE}.


\begin{algorithm}
  \caption{Hedge}
  \label{alg:HEDGE}
  \begin{algorithmic}
    \REQUIRE $\gamma > 0$
    \REQUIRE $G_0(i) \leftarrow 0$ \quad\textbf{foreach} $i \in \mathcal{A}$
    \FOR{$t = 1, 2, \ldots$}
    \FOR{$i \in \mathcal{A}$}
    \STATE $p_t(i) \leftarrow \frac{\exp{\gamma G_{t-1}(i)}}{\sum_{j
        \in \mathcal{A}}\exp{\gamma G_{t-1}(j)}}$
    \ENDFOR
    \STATE $\vec{a}_t \sim (p_t(1), \ldots, p_t(|\mathcal{A}|))$
    \COMMENT{sample an action from the distribution $(p_t(k))$}
    \FOR{$i \in \mathcal{A}$}
    \STATE $r_t(i) \leftarrow r_t(i|\vec{b}_t)$
    \STATE $G_t(i) \leftarrow G_{t-1}(i) + r_t(i)$
    \ENDFOR
    \ENDFOR
  \end{algorithmic}
\end{algorithm}

\subsection{EXP3}

EXP3~\citep{auer2001nonstochastic} is a generalization of Hedge to the
partial information setting.  In order to maintain estimates for the
importance weights, Hedge requires reward information for each
possible action at each time step.  EXP3 works by wrapping Hedge in an
outer loop which simulates a fully observed reward vector at each time
step.  EXP3 selects actions based on a mixture of the policy found by
Hedge and a uniform distribution.  EXP3 is able to function in the
presence of partial information, but this comes at the cost of
substantially worse theoretical guarantees.  Pseudo code for EXP3 is
shown in Algorithm~\ref{alg:exp3}.

\begin{algorithm}
  \caption{EXP3}
  \label{alg:exp3}
  \begin{algorithmic}
    \REQUIRE $\gamma \in (0, 1]$
    \STATE Initialize \textbf{Hedge}$(\gamma)$
    \FOR{$t \in 1, 2, \ldots$}
    \STATE Receive $\vec{p}_t$ from \textbf{Hedge}
    \STATE $\hat{\vec{p}}_t \leftarrow (1-\gamma)\vec{p}_t +
    \frac{\gamma}{|\mathcal{A}|}$
    \STATE $\vec{a}_t \sim (\hat{p}_t(1), \ldots,
    \hat{p}_t(|\mathcal{A}|))$
    \STATE Simulate reward vector for \textbf{Hedge} where
    $\hat{r}_t(j) \leftarrow \begin{cases} r_t(j)/p_t(j) & \mathrm{if}
      j = \vec{a}_t \\ 0 & \mathrm{otherwise} \end{cases}$
    \ENDFOR
  \end{algorithmic}
\end{algorithm}

\subsection{Bayesian Optimization}

Both Hedge and EXP3 discretize the space of possible fixation points
and learn a distribution over this finite set.  In contrast, Bayesian
optimization is able to treat the space of possible fixation points as
fully continuous by placing a smoothness prior on how reward is
expected to vary with location.  Intuitively, if we know the reward at
one location, then we expect other, nearby locations to produce
similar rewards.  Gaussian Process priors encode this type of
belief~\citep{rasmussen2006gaussian}, and have been used extensively
for optimization of cost functions when it is important to minimize
the total number of function evaluations~\citep{brochu2010tutorial}.

We model the latent reward function $r_t(\vec{a}_t|\vec{b}_t)
\triangleq r(\vec{a}_t|\vec{b}_t,\vec{\theta}_t)$ as a zero mean
Gaussian Process
\begin{align*}
  r(\vec{a}_t|\vec{b}_t,\vec{\theta}_t) \sim \mathcal{GP}(\vec{0},
  k(\vec{a}_t,\vec{a}_t^\prime| \vec{b}_t, \vec{\theta}_t)) \enspace,
\end{align*}
where $\vec{b}_t$ is the belief state (see
Section~\ref{sec:state-space-model}), and $\vec{\theta}_t$ are the
model hyperparameters.  The kernel function $k(\cdot,\cdot)$, gives
the covariance between the reward at any two gaze locations.  To ease
the notation, the explicit dependence of $r(\cdot)$ and
$k(\cdot,\cdot)$ on $\vec{b}_t$ and $\vec{\theta}_t$ will be dropped.

We assume that the true reward function $r(\cdot)$ is not directly
measurable, and what we observe are measurements of this function
corrupted by Gaussian noise.  That is, at each time step the
instantaneous reward $r_t$, is given by
\begin{align*}
  r_t = r(\vec{a}_t) + \sigma_n\delta_n \enspace,
\end{align*}
where $\delta_n \sim \mathcal{N}(0, 1)$ and $\sigma_n$ is a
hyperparameter indicating the amount of observation noise, which we
absorb into $\vec{\theta}_t$.

Given a set of observations we can compute the posterior predictive
distribution for $r(\cdot)$:
\begin{align}
  r(\vec{a}|\vec{r}_{1:t}, \vec{a}_{1:t}) &\sim \mathcal{N}(m_{t}(\vec{a}),
  s^2_{t}(\vec{a})) \enspace, \label{eq:posterior}\\
  m_{t}(\vec{a}) &= \vec{k}^\T[\vec{K} +
  \sigma_n^2\vec{I}]^{-1}\vec{r}_{1:t} \enspace, \nonumber\\
  s^2_{t}(\vec{a}) &= k(\vec{a}, \vec{a}) - \vec{k}^\T[\vec{K} +
  \sigma_n^2\vec{I}]^{-1}\vec{k} \enspace, \nonumber
\end{align}
where
\begin{align*}
  \vec{K} &= \begin{bmatrix} k(\vec{a}_1, \vec{a}_1) & \cdots &
    k(\vec{a}_1, \vec{a}_t) \\
    \vdots & \ddots & \vdots \\
    k(\vec{a}_t, \vec{a}_1) & \cdots & k(\vec{a}_t, \vec{a}_t)
  \end{bmatrix}
  \enspace, \\
  \vec{k} &= \begin{bmatrix}
    k(\vec{a}_{1}, \vec{a}) & \cdots & k(\vec{a}_{t},
    \vec{a})
  \end{bmatrix}^\T
  \enspace, \\
  \vec{r}_{1:t} &= \begin{bmatrix}
    r_1 & \cdots & r_t
  \end{bmatrix}^\T
  \enspace.
\end{align*}

It remains to specify the form of the kernel function, $k(\cdot,
\cdot)$.  We experimented with several possibilities, but found that
the specific form of the kernel function is not critical to the
performance of this method.  For the experiments in this paper we used
the squared exponential kernel,
\begin{align*}
  k(\vec{a}_i, \vec{a}_j) &= \sigma_m^2
  \exp{-\frac{1}{2}\sum_{k=1}^D\left(\frac{a_{i,k}-a_{j,k}}{\ell_k}\right)^2} \enspace,
\end{align*}
where $\sigma_m^2$ and the $\{\ell_1, \ldots, \ell_D\}$ are
hyperparameters.

Equation~\ref{eq:posterior} is a Gaussian Process estimate of the
reward surface and can be used to select a fixation point for the next
time step.  This estimate gives both a predicted reward value and an
associated uncertainty for each possible fixation point.  This is the
strength of Gaussian Processes for this type of optimization problem,
since the predictions can be used to balance exploration (choosing a
fixation point where the reward is highly uncertain) and exploitation
(choosing a point we are confident will have high reward).

There are many selection methods available in the literature which
offer different tradeoffs between these two criteria.  In this paper
we use GP-UCB~\citep{srinivas2009gaussian} which selects
\begin{align}
  \label{eq:ucb}
  \vec{a}_{t+1} = \arg\max_{\vec{a}} m_t(\vec{a}) +
  \sqrt{\beta_t}s_t(\vec{a})
\end{align}
where $\beta_t$ is a parameter.  The setting $\beta_t =
2\log(t^3\pi^2/3\delta)$ (with $\delta=0.001$) is used throughout this
paper.

Equation~\ref{eq:ucb} must still be optimized to find $\vec{a}_{t+1}$,
which can be performed using standard global optimization tools.  We
use \texttt{DIRECT}~\citep{jones1993lipschitzian} due to the existence
of a readily available implementation.

\begin{figure}[tbh]
  \centering
  \includegraphics[width=0.4\linewidth]{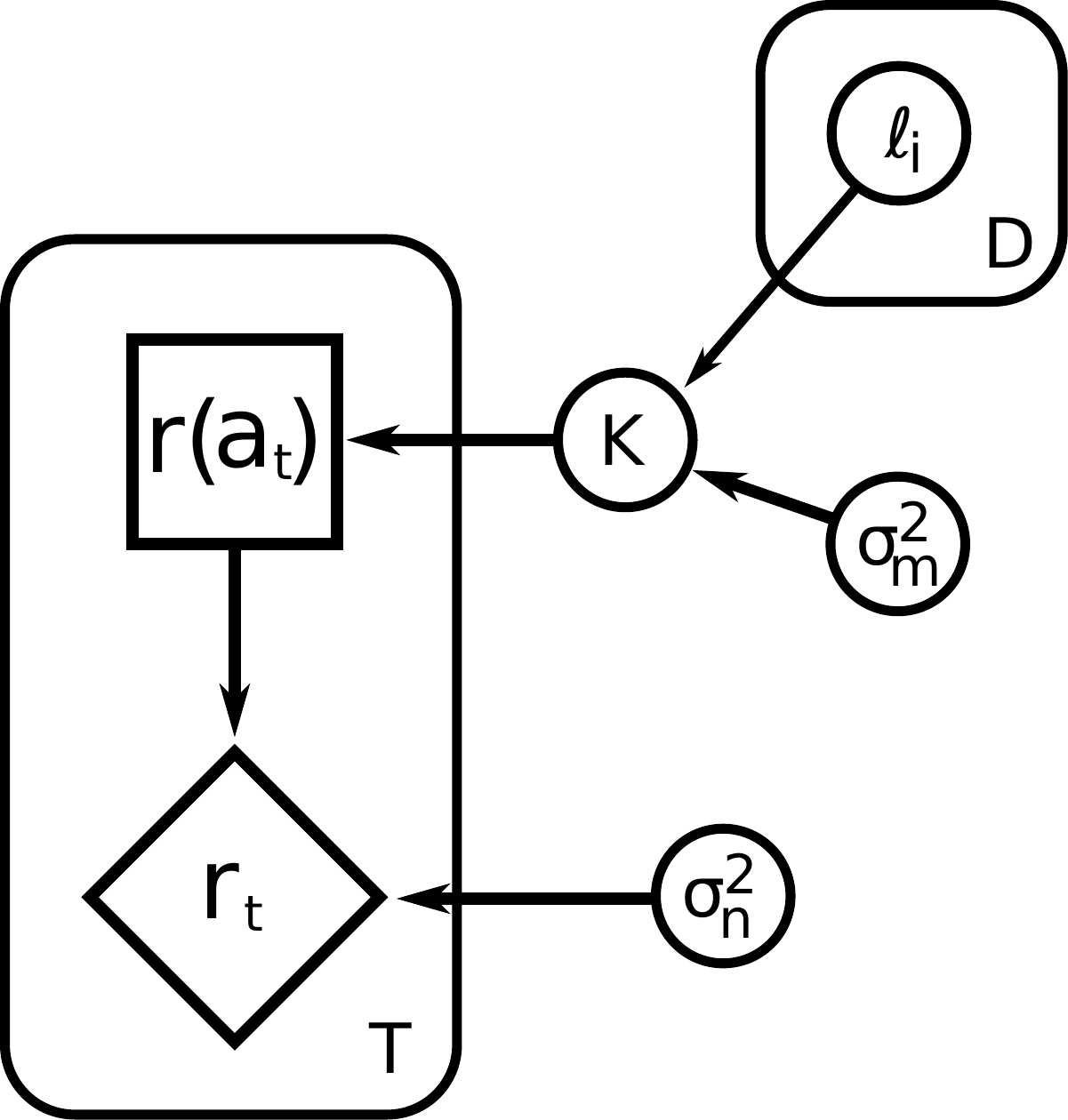}
  \caption{\capstyle{Graphical model for Bayesian optimization. The
      $\ell_i$ are length scales in each dimension, $\sigma_m^2$ is
      the magnitude parameter and $\sigma_n^2$ is the noise level. In
      our model $\sigma_m^2$ and $\sigma_n^2$ follow a uniform prior
      and the $\ell_i$ follow independent Student-t priors.}}
  \label{fig:bayesoptgm}
\end{figure}

The Gaussian Process regression is controlled by several
hyperparameters (see Figure~\ref{fig:bayesoptgm}): $\sigma_m^2$
controls the overall magnitude of the covariance, and $\sigma_n^2$
controls the amount of observation noise.  The remaining parameters
$\{\ell_1,\ldots,\ell_D\}$ are length scale parameters which control
the range of the covariance effects in each dimension.

Treatment of the hyperparameters requires special consideration in
this setting.  The pure Bayesian approach is to put a prior on each
parameter and integrate them out of the predictive distribution.
However, since the integrals involved are not tractable analytically,
this requires computationally expensive numerical approximations.
Speed is an issue here since GP-UCB requires that we optimize a
function of the posterior process at each time step so, for instance,
computing Monte Carlo averages for each evaluation of
Equation~\ref{eq:posterior} is prohibitively slow.

An alternative approach is to choose parameter values via maximum
likelihood.  This can be done quickly, and allows us to make speedy
predictions; however, in this case we suffer from problems of data
scarcity, particularly early in the tracking process when few
observations have been made.  The length scale parameters are
particularly prone to receiving very poor estimates when there is
little data available.

We have found that using informative priors for the length scale
parameters and making MAP, rather than ML, estimates at each time step
provides a solution to the problems described above.  MAP estimates
can be made quickly using gradient optimization
methods~\citep{rasmussen2006gaussian}, and informative priors provide
resistance to the problems encountered with ML.  The experiments in
Section~\ref{sec:experiments} place uniform priors on the magnitude
and noise parameters and place independent Student-t priors on each
length scale parameter.  The experiments also use an initial data
collection phase of 10 time steps before any adjustment of the
parameters is made.


%% file: algorithm.tex
\section{Algorithm}


Since the belief state cannot be computed analytically, we will adopt
particle filtering to approximate it.  The full algorithm is shown in
Algorithm~\ref{alg:full}.

\begin{algorithm}
  \begin{algorithmic}
    \ALGORITHMSECTION{Initialization}
    \FOR{$i=1$ \textbf{to} $N$}
    \STATE $\vec{x}_0^{(i)} \sim p(\vec{x}_0)$
    \ENDFOR
    \STATE Initialize the policy $\pi_1(\cdot)$
    \COMMENT{How this is done depends on the control strategy}
    \ENDALGORITHMSECTION
    \FOR{$t=1\ldots$\ }
    \ALGORITHMSECTION{Importance sampling}
    \FOR[Predict the next state]{$i=1$ \textbf{to} $N$}
    \STATE $\widetilde{\mathbf{x}}_{t}^{\left( i\right) }\sim
    q_{t}\left( \left. d{\xv }_{t}^{(i)}\right|
      \widetilde{\xv}_{0:t-1}^{(i)},\hv_{1:t},\av_{1:t}\right)$
    \STATE $\widetilde{\vec{x}}_{0:t}^{(i)} \leftarrow
    \left(\vec{x}_{0:t-1}^{(i)}, \widetilde{\vec{x}}_{t}^{(i)}\right)$
    \ENDFOR
    \STATE $k^\star \sim \pi_t(\cdot)$
    \COMMENT{Select an action according to the policy}
    \FOR[Evaluate the importance weights]{$i=1$ \textbf{to} $N$}
    \STATE 
    \vspace{-0.7cm}
    \begin{align*}
    \widetilde{w}_{t}^{\left( i\right) } \leftarrow
    \frac{p\left(\hv_t|\widetilde{\xv}_{t}^{(i)},\av_{t}=k^\star \right)
      p\left(\widetilde{\xv }_{t}^{(i)}| \widetilde{\xv}_{0:t-1}^{(i)},\av_{t-1}\right)}
    {q_{t}\left( \left. \widetilde{\xv }_{t}^{(i)}\right| \widetilde{\xv }%
        _{0:t-1}^{(i)},\hv_{1:t},\av_{1:t}\right)}
    & \hspace{5cm}
    \end{align*}
    \ENDFOR
    \FOR[Normalize the importance weights]{$i=1$ \textbf{to} $N$}
    \STATE $w_t^{(i)} \leftarrow \frac{\widetilde{w}_t^{(i)}}{\sum_{j=1}^{N}
      \widetilde{w}_t^{(j)}}$
    \ENDFOR
    \ENDALGORITHMSECTION
    \ALGORITHMSECTION{Gaze control}
    \STATE $r_t = \sum_{i=1}^N (w_t^{(i)})^2$
    \COMMENT{Receive reward for the chosen action}
    \STATE Incorporate $r_t$ into the policy to create
    $\pi_{t+1}(\cdot)$
    \ENDALGORITHMSECTION
    \ALGORITHMSECTION{Selection}
    \STATE Resample with replacement $N$ particles $\left(
      \mathbf{x}_{0:t}^{\left( i\right) };i=1,\ldots ,N\right)$ from
    the set $\left( \widetilde{\mathbf{x}}_{0:t}^{\left( i\right)
      };i=1,\ldots,N\right)$ according to the normalized importance
    weights ${w}_{t}^{\left( i\right)}$
    \ENDALGORITHMSECTION
    \ENDFOR
  \end{algorithmic}
  \caption{Particle filtering algorithm with gaze control.  The
    algorithm shown here is for partial information strategies.  For
    full information strategies the importance sampling step is done
    independently for each possible action and the gaze control step
    is able to use reward information from each possible action to
    create the new strategy $\pi_{t+1}(\cdot)$. \vspace{2mm}}
  \label{alg:full}
\end{algorithm}

We refer readers to \cite{315} for a more in depth treatment of these
sequential Monte Carlo methods.  Assume that at time $t-1$ we have
$N\gg 1$ particles (samples) $\{ \xv_{0:t-1}^{\left( i\right) }\}
_{i=1}^{N}$ distributed according to $ p\left( d\xv
  _{0:t-1}|\hv_{1:t-1},\av_{1:t-1}\right) $. We can approximate this
belief state with the following empirical distribution
\begin{align*}
  \widehat{p}\left( d\xv _{0:t-1}|\hv_{1:t-1},\av_{1:t-1}\right)
  \triangleq \frac{1}{N} \sum_{i=1}^{N}\delta _{\xv _{0:t-1}^{\left(
        i\right) }}\left( d\xv _{0:t-1}\right) \enspace.
\end{align*}
Particle filters combine sequential importance sampling with a
selection scheme designed to obtain $N$ new particles $\{ \xv_{0:t}^{
  \left( i\right)}\}_{i=1}^{N}$ distributed approximately according to
$p\left( d\xv_{0:t}|\hv_{1:t},\av_{1:t}\right)$.

\subsection{Importance sampling step}

The joint distributions $p\left( d\xv
  _{0:t-1}|\hv_{1:t-1},\av_{1:t-1}\right)$ and $p\left(
  d\xv_{0:t}|\hv_{1:t},\av_{1:t}\right)$ are of different
dimension. We first modify and extend the current paths $\xv
_{0:t-1}^{\left( i\right) }$ to obtain new paths $\widetilde{\xv
}_{0:t}^{\left( i\right) }$ using a proposal kernel
$q_{t}\left(d\widetilde{\xv }_{0:t}|\xv
  _{0:t-1},\hv_{1:t},\av_{1:t}\right).$ As our goal is to design a
sequential procedure, we set
\begin{align*}
  q_{t}\left( \left. d\widetilde{\xv }_{0:t}\right| \xv
    _{0:t-1},\hv_{1:t},\av_{1:t}\right) =\delta _{\xv _{0:t-1}}\left(
    d\widetilde{\xv }_{0:t-1}\right) q_{t}\left(
    \left. d\widetilde{\xv }_{t}\right| \widetilde{\xv }%
    _{0:t-1},\hv_{1:t},\av_{1:t}\right) \enspace,
\end{align*}
that is $\widetilde{\xv }_{0:t}=\left( \xv _{0:t-1},%
  \widetilde{\xv }_{t}\right) $. The aim of this kernel is to obtain
new paths whose distribution
\begin{align*}
  q_{t}\left( d\widetilde{\xv }_{0:t}|\hv_{1:t},\av_{1:t}\right)
  =p\left( d\widetilde{\xv }_{0:t-1}|\hv_{1:t-1},\av_{1:t-1}\right)
  q_{t}\left( \left. d%
      \widetilde{\xv }_{t}\right| \widetilde{\xv
    }_{0:t-1},\hv_{1:t},\av_{1:t}\right)
  \enspace,
\end{align*}
is as ``close'' as possible to $p\left( d\widetilde{\xv
  }_{0:t}|\hv_{1:t},\av_{1:t}\right)$.  Since we cannot choose
$q_{t}\left( d\widetilde{\xv }%
  _{0:t}|\hv_{1:t},\av_{1:t}\right) =p\left( d\widetilde{\xv
  }_{0:t}|\hv_{1:t},\av_{1:t}\right) $ because this is the quantity we
are trying to approximate in the first place, it is necessary to
weight the new particles so as to obtain consistent estimates. We
perform this ``correction'' with importance sampling, using the
weights
\begin{equation}
\widetilde{w}_t 
=
\widetilde{w}_{t-1} \frac{p\left(\hv_t|\widetilde{\xv}_{t},\av_{t}\right)
  p\left(d\widetilde{\xv }_{t}| \widetilde{\xv}_{0:t-1},\av_{t-1}\right)}
{q_{t}\left( \left. d\widetilde{\xv }_{t}\right| \widetilde{\xv }%
    _{0:t-1},\hv_{1:t},\av_{1:t}\right)}.
\nonumber \label{eq:impweight}
\end{equation}

The choice of the transition prior as proposal distribution is by far
the most common one. In this case, the importance weights reduce to
the expression for the likelihood. However, it is possible to
construct better proposal distributions, which make use of more recent
observations, using object detectors~\citep{Okuma:2004}, saliency
maps~\citep{Itti1998}, optical flow, and approximate filtering methods
such as the unscented particle filter. One could also easily
incorporate strategies to manage data association and other tracking
related issues.  After normalizing the weights, $w_t^{(i)} =
\frac{\widetilde{w}_t^{(i)}}{\sum_{j=1}^{N} \widetilde{w}_t^{(j)}}$,
we obtain the following estimate of the filtering distribution:
\begin{equation}
\widetilde{p}\left( d\xv _{0:t}|\hv_{1:t},\av_{1:t}\right)
=\sum_{i=1}^{N}w_{t}^{\left( i\right) }\delta _{\widetilde{\xv }%
_{0:t}^{\left( i\right) }}\left( d\xv _{0:t}\right).
\nonumber \label{eq:meaweight}
\end{equation}

Finally a selection step is used to obtain an ``unweighted''
approximate empirical distribution
$\hat{p}(d\vec{x}_{0:t}|\vec{h}_{1:t}, \vec{a}_{1:t})$ of the weighted
measure $\tilde{p}(d\vec{x}_{0:t}|\vec{h}_{1:t}, \vec{a}_{1:t})$.  The
basic idea is to discard samples with small weights and multiply those
with large weights.  The use of a selection step is key to making the
SMC procedure effective; see~\cite{315} for details on how to
implement this black box routine.


%% file: experiments.tex
\section{Experiments}
\label{sec:experiments}

\subsection{Full Information Policies}

In this section, three experiments are carried out to evaluate
quantitatively and qualitatively the proposed approach. The first
experiment provides comparisons to other control policies on a
synthetic dataset. The second experiment, on a similar synthetic
dataset, demonstrates how the approach can handle large variations in
scale, occlusion and multiple targets. The final experiment is a
demonstration of tracking and classification performance on several
real videos. For the synthetic digit videos, we trained the
first-layer RBMs on the foveated images, while for the real videos we
trained factored-RBMs on foveated natural image
patches~\citep{Ranzato2010}.

\begin{figure*}[t!]
 \begin{center}
   \includegraphics[width=\textwidth]{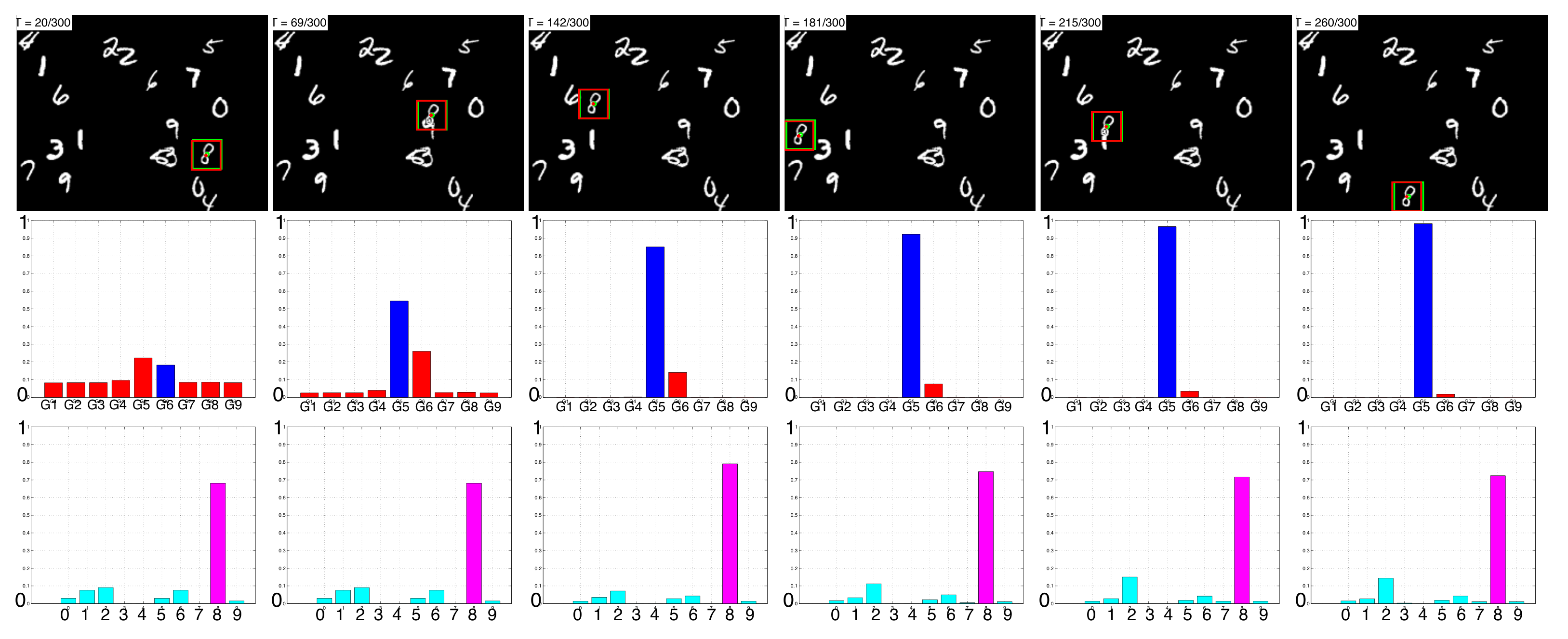}
 \end{center}
 \vspace{-10pt}
 \caption{\capstyle{Tracking and classification accuracy results with
     the learned policy. \textbf{First row:} position of the target
     and estimate over time. \textbf{Second row:} policy distribution
     over the 9 gazes; hedge clearly converges to the most reasonable
     policy. \textbf{Third row:} cumulative class distribution for
     recognition.}}
 \label{fig:res_learned}
\end{figure*}

The first experiment uses 10 video sequences (one for each digit)
built from the MNIST dataset. Each sequence contains a moving digit
and static digits in the background (to create distractions). The
objective is to track and recognize the moving digit; see
Figure~\ref{fig:res_learned}.  The gaze template had $K=9$ gaze
positions, chosen so that gaze G5 was at the center. The location of
the template was initialized with optical flow.

We compare the Hedge learning algorithm against algorithms with
deterministic and random policies. The deterministic policy chooses
each gaze in sequence and in a particular pre-specified order, whereas
the random policy selects a gaze uniformly at random.  We adopted the
Bhattacharyya distance in the specification of the observation
model. A multi-fixation RBM was trained to map the first layer hidden
units of three time consecutive time steps into a second hidden layer,
and trained a logistic regressor to further map to the 10 digit
classes. We used the transition prior as proposal for the particle
filter.

\input{tables/tracking-error-full-info}
\input{tables/classification-full-info}

Tables~\ref{tab:1} and \ref{tab:2} report the comparison
results. Tracking accuracy was measured in terms of the mean and
standard deviation (in brackets) over time of the distance between the
target ground truth and the estimate; measured in pixels.  The
analysis highlights that the error of the learned policy is always
below the error of the other policies. In most of the experiments, the
tracker fails when an occlusion occurs for the deterministic and the
random policies, while the learned policy is successful. This is very
clear in the videos at: 

The loss of track for the simple policies is mirrored by the high
variance results in Table~\ref{tab:1} (experiments $0$, $1$, $4$, and
so on). The average mean and standard deviations (last column of
Table~\ref{tab:1}) make it clear that the proposed strategy for
learning a gaze policy can be of enormous benefit. The improvements in
tracking performance are mirrored by improvements in classification
performance (Table~\ref{tab:2}).

Figure~\ref{fig:res_learned} provides further anecdotal evidence for
the policy learning algorithm.  The top sequence shows the target and
the particle filter estimate of its location over time. The middle
sequence illustrates how the policy changes over time. In particular,
it demonstrates that hedge can effectively learn where to look in
order to improve tracking performance (we chose this simple example as
in this case it is obvious that the center of the eight (G5) is the
most reliable gaze action). The classification results over time are
shown in the third row.

\begin{figure*}[t!]
 \begin{center}
 \vspace{-1cm}
   \includegraphics[width=\textwidth]{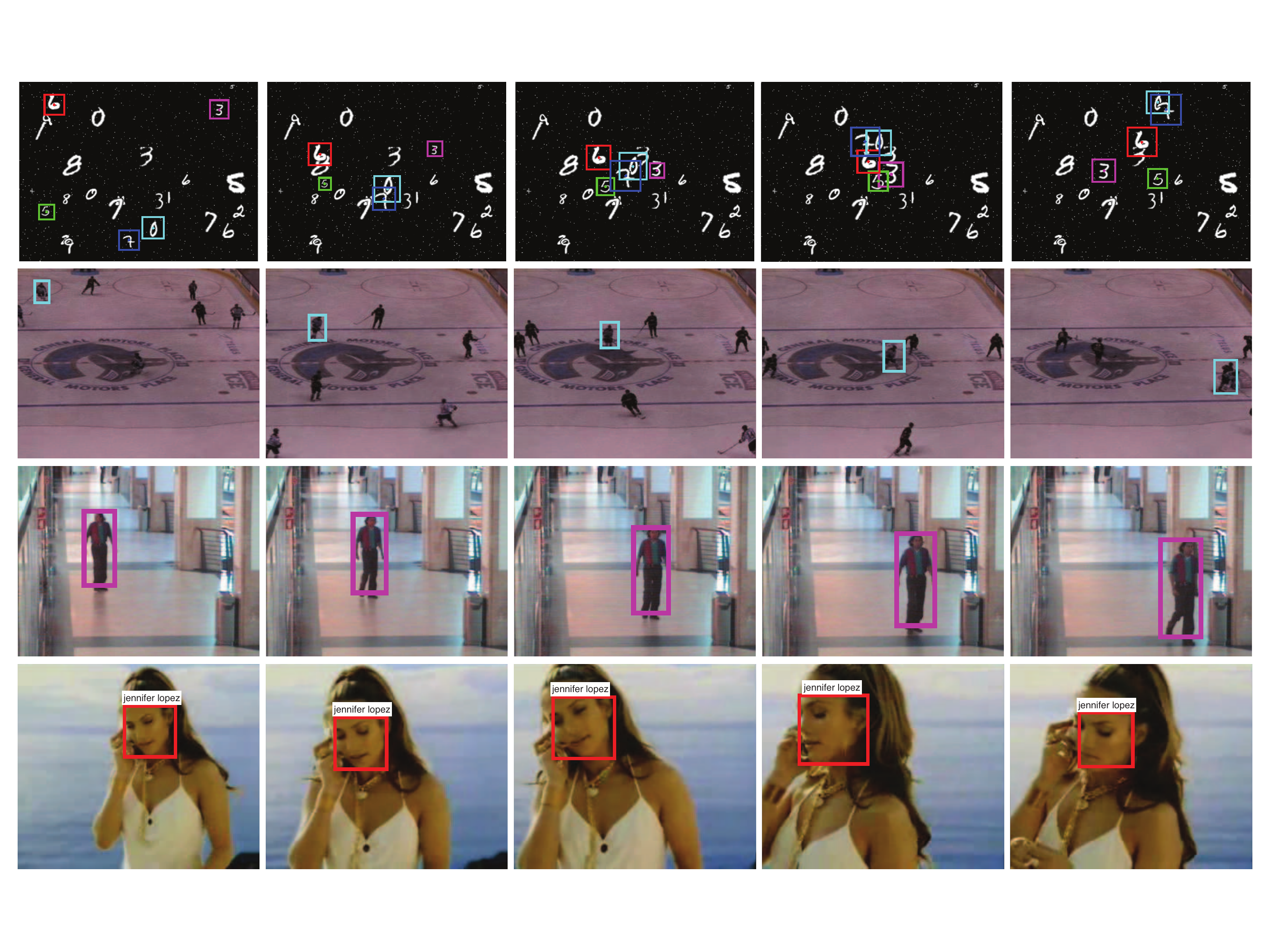}
 \end{center}
 \vspace{-1.5cm}
 \caption{\capstyle{\textbf{Top:} Multi-target tracking with
     occlusions and changes in scale on a synthetic
     video. \textbf{Middle and bottom:} Tracking in real video
     sequences.}}
 \label{fig:hockey}
\end{figure*}

The second experiment addresses a similar video sequence, but tracking
multiple targets. The image scale of each target changes significantly
over time, so the algorithm has to be invariant with respect to these
scale transformations.  In this case, we used a mixture proposal
distribution consisting of motion detectors and the transition
prior. We also tested a saliency proposal but found it to be less
effective than the motion detectors for this
dataset. Figure~\ref{fig:hockey} (top) shows some of the video frames
and tracks. The videos allow one to better appreciate the performance
of the multi-target tracking algorithm in the presence of occlusions.

Tracking and classification results for the real videos are shown in
Figure~\ref{fig:hockey} and the accompanying videos.

\subsection{Partial Information Policies}

In this section, two experiments are carried out to evaluate the
performance of the different gaze selection policies.  

In the first experiment we compare the performance of each gaze
selection method on a data set of several videos of digits from the
MNIST data set moving on a black background.  The target in each video
encounters one or more partial occlusions which the tracking algorithm
must handle gracefully.  Additionally, each video sequence has been
corrupted with 30\% noise.  We measure the error between the estimated
track and the ground truth for each gaze selection method, and
demonstrate that Bayesian optimization preforms comparably to Hedge,
but that EXP3 is not able to reach a satisfactory level of
performance.  We also demonstrate qualitatively that the Bayesian
optimization approach learns good gaze selection policies on this data
set.

Our second experiment provides evidence that the Bayesian optimization
method can generalize to real world data.

\input{tables/tracking-error-partial-info}

Table~\ref{tab:experiment2} reports the results from our first
experiment.  The table shows the mean tracking error, measured by
averaging distance between the estimated and ground truth track over
the entire video sequence.  Here we see that the Bayesian optimization
approach compares favorably to Hedge in terms of tracking performance,
and that EXP3 preforms substantially worse than the other two methods.
Although Hedge preforms marginally better than Bayesian optimization,
it is important to remember that Bayesian optimization solves a
significantly more difficult problem.  Hedge relies on discretizing
the action space, and must have access to the rewards for all possible
actions at each time step.  In contrast, Bayesian optimization
considers a fully continuous action space, and receives reward
information only for the chosen actions.

\begin{figure}[tbh]
  \centering
  \includegraphics[width=\linewidth]{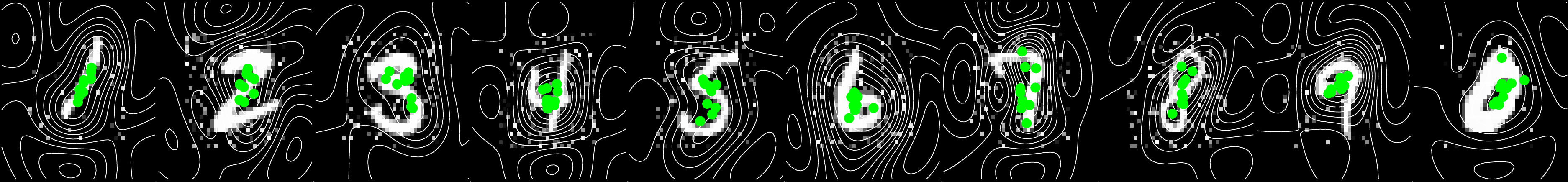} \\
  \includegraphics[width=\linewidth]{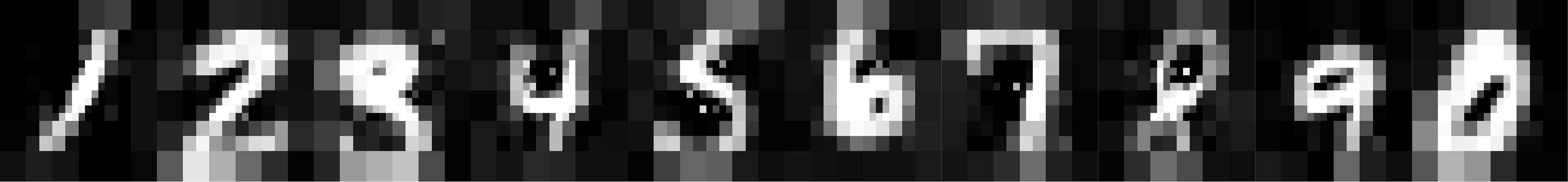}
  \caption{\capstyle{\textbf{Top:} Digit templates with the estimated
      reward surfaces superimposed.  Markers indicate the best
      fixation point found in each of ten runs.  \textbf{Bottom:} A
      visualization of the image found by averaging the best fixation
      points found across ten runs.}}
  \label{fig:digitBanner}
\end{figure}

Figure~\ref{fig:digitBanner} shows the reward surfaces learned for
each digit by Bayesian optimization, as well as a visualization of the
overall best fixation points using data aggregated across ten runs.
The optimal fixation points found by the algorithm are tightly
clustered, and the resulting observations are very distinguishable.

\begin{figure}[tbh]
  \centering
  \includegraphics[width=0.268\linewidth]{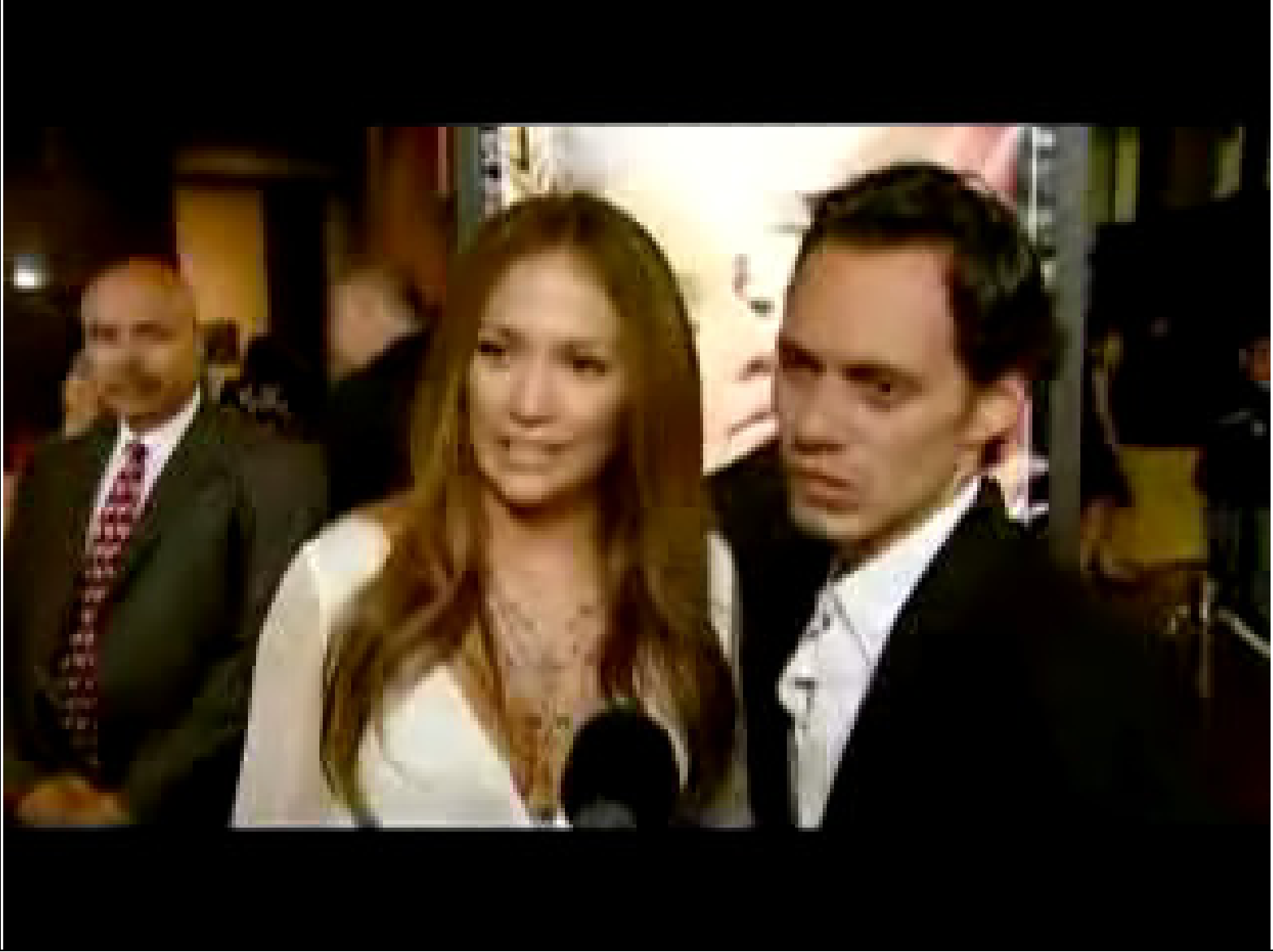}
  \quad
  \includegraphics[width=0.2\linewidth]{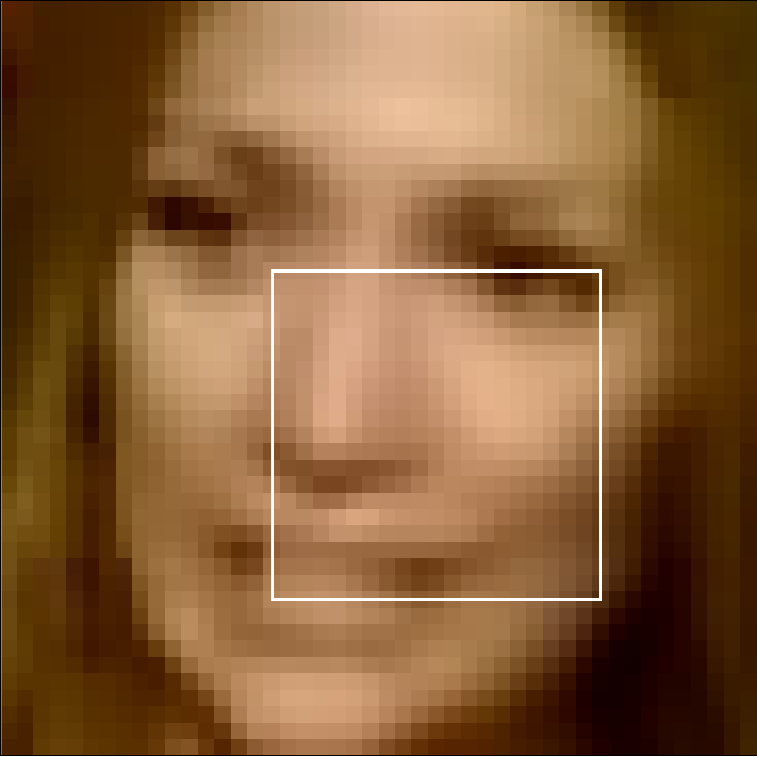}
  \quad
  \includegraphics[width=0.2\linewidth]{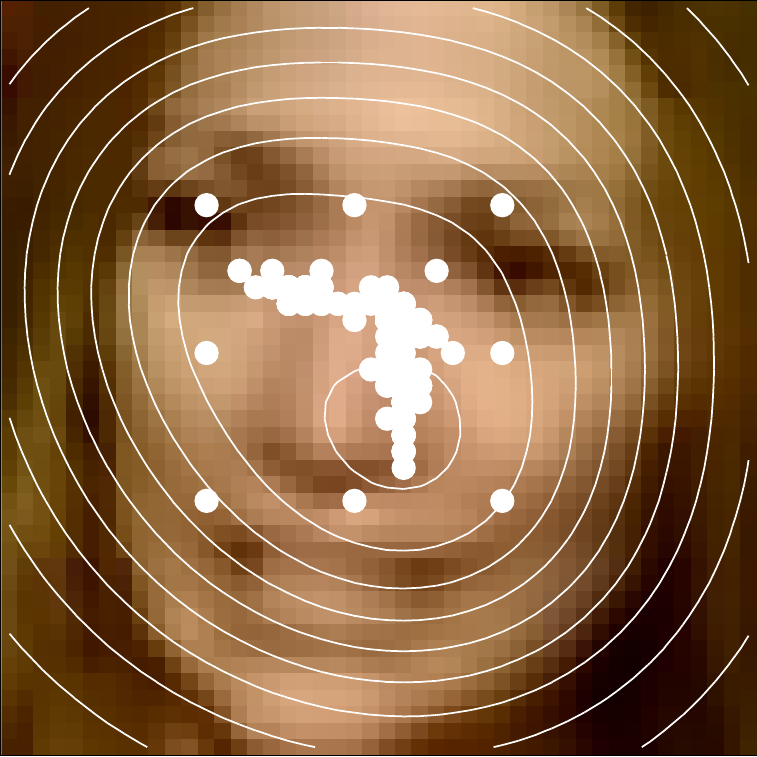}
  \quad
  \includegraphics[width=0.2\linewidth]{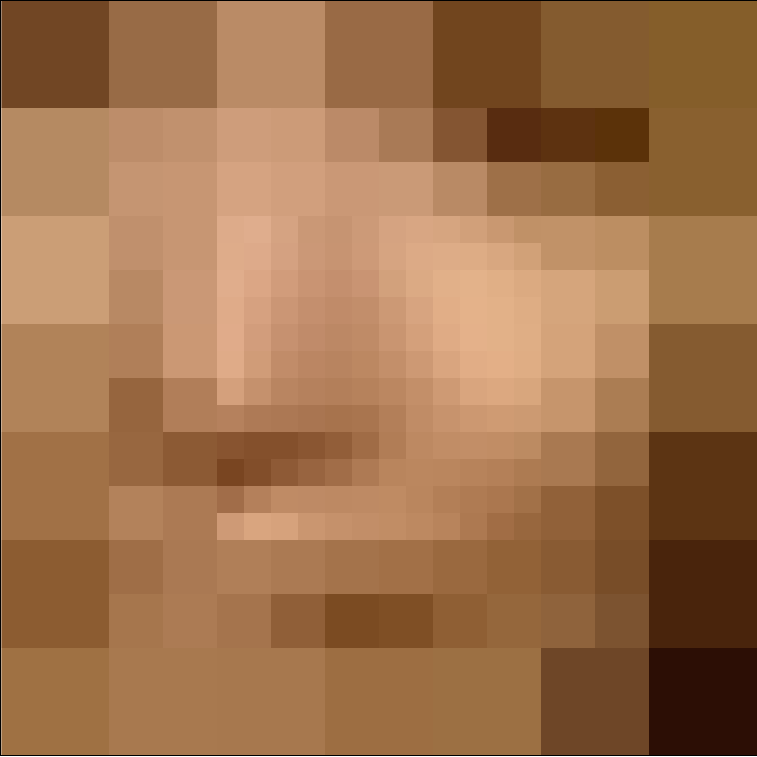}
  \caption{\capstyle{Results on a real data set.  \textbf{Far left:}
      An example frame from the video sequence. \textbf{Center left:}
      The tracking template with the optimal fixation window
      highlighted.  \textbf{Center right:} The reward surface produced
      by Bayesian optimization.  The white markers show the centers of
      each fixation point in a single tracking run. \textbf{Right:}
      Input to the observation model when fixating on the best
      point. (Best viewed from a distance).}}
  \label{fig:face}
\end{figure}

In our second experiment we use the Youtube celebrity dataset
from~\cite{kim2008face}.  This data set consists of several videos of
celebrities taken from Youtube and is challenging for tracking
algorithms as the videos exhibit a wide variety of illuminations,
expressions and face orientations.  We run our tracking model using
Bayesian optimization to learn a gaze selection policy on this data
set, and present some results in Figure~\ref{fig:face}.  Although we
report only qualitative results from this experiment, it provides
anecdotal evidence that Bayesian optimization is able to form a good
gaze selection policy on real world data.

%% file: tables/tracking-error-full-info.tex
\begin{table*}[t!]
\label{tab:1}
\begin{center}
\begin{tiny}
\begin{sc}
\begin{tabular}{|p{1.3cm}|p{0.7cm}|p{0.7cm}|p{0.7cm}|p{0.7cm}|p{0.7cm}|p{0.7cm}|p{0.7cm}|p{0.7cm}|p{0.7cm}|p{0.7cm}||p{0.7cm}|}
\hline
 & 0 & 1 & 2 & 3 & 4 & 5 & 6 & 7 & 8 & 9 & Avg.\\
 \hline
 Learned policy  & \textbf{1.2} (1.2) & \textbf{3.0} (2.0) & \textbf{2.9} (1.0) & \textbf{2.2} (0.7) & \textbf{1.0} (1.9) & \textbf{1.8} (1.9) & \textbf{3.8} (1.0) & \textbf{3.8} (1.5) & \textbf{1.5} (1.7) & \textbf{3.8} (2.8) & \textbf{2.5} (1.6) \\
\hline
 Deterministic policy & 18.2 (29.6) & 536.9 (395.6) & 104.4 (69.7) & 2.9 (2.2) & 201.3 (113.4) & 4.6 (4.0) & 5.6 (3.1) & 64.4 (45.3) & 142.0 (198.8) & 144.6 (157.7) & 122.5 (101.9) \\
 \hline
 Random policy & 41.5 (54.0) & 410.7 (329.4) & 3.2 (2.0) & 3.3 (2.4) & 42.8 (60.9) & 6.5 (9.6) & 5.7 (3.2) & 80.7 (48.6) & 38.9 (50.6) & 225.2 (241.6) & 85.9 (80.2) \\
 \hline
\end{tabular}
\end{sc}
\end{tiny}
\end{center}
\caption{\capstyle{Tracking error (in pixels) on several video sequences using different policies for gaze selection.}}
\end{table*}

%% file: tables/classification-full-info.tex
\begin{table*}[t!]
\begin{center}
\begin{tiny}
\begin{sc}
\begin{tabular}{|p{1.3cm}|p{0.7cm}|p{0.9cm}|p{0.7cm}|p{0.7cm}|p{0.7cm}|p{0.9cm}|p{0.9cm}|p{0.7cm}|p{0.7cm}|p{0.7cm}||p{0.7cm}|}
\hline
 & 0 & 1 & 2 & 3 & 4 & 5 & 6 & 7 & 8 & 9 & Avg.\\
 \hline
 Learned policy & 95.62\% & \textbf{100.00}\% & \textbf{99.66}\% & 99.33\% & \textbf{99.66}\% & \textbf{100.00}\% & \textbf{100.00}\% & \textbf{98.32}\% & \textbf{97.98}\% & \textbf{89.56}\% & \textbf{98.01}\% \\
\hline
 Deterministic policy & \textbf{99.33}\% & \textbf{100.00}\% & 98.99\% & 94.95\% & 5.39\% & 98.32\% & 0.00\% & 29.63\% & 52.19\% & 0.00\% & 57.88\% \\
\hline
 Random policy & 98.32\% & \textbf{100.00}\% & 96.30\% & \textbf{99.66}\% & 29.97\% & 96.30\% & 89.56\% & 22.90\% & 12.79\% & 13.80\% & 65.96\% \\
\hline
 \end{tabular}
\end{sc}
\end{tiny}
\end{center}
\label{tab:2}
\vspace{-0.2cm}
\caption{\capstyle{Classification accuracy on several video sequences using different policies for gaze selection.}}
\end{table*}

%% file: tables/tracking-error-partial-info.tex
\begin{table}[t]
  \centering
  \begin{tiny}

\begin{tabularx}{\linewidth}{|X|X|X|X|X|X|X|X|X|X|X||X|}
\hline
~ & \textbf{0} & \textbf{1} & \textbf{2} & \textbf{3} & \textbf{4} & \textbf{5} & \textbf{6} & \textbf{7} & \textbf{8} & \textbf{9} & \textbf{Avg} \\ 
\hline
\textbf{Bayesopt} & 5.36 \newline (2.32) & 7.92 \newline (2.52) & 2.62 \newline (3.89) & 4.05 \newline (1.67) & 1.70 \newline (5.10) & 8.31 \newline (3.35) & 4.94 \newline (2.28) & 12.09 \newline (3.53) & 1.52 \newline (2.76) & 9.06 \newline (1.66) & \textbf{5.76 \newline (2.91)}\\ 
\hline
\textbf{Hedge} & 2.97 \newline (1.56) & 3.20 \newline (2.19) & 2.97 \newline (1.99) & 2.92 \newline (2.00) & 3.14 \newline (1.80) & 2.96 \newline (2.08) & 2.86 \newline (1.96) & 2.98 \newline (1.76) & 2.81 \newline (1.64) & 3.15 \newline (3.73) & \textbf{3.00 \newline (2.07)}\\ 
\hline
\textbf{EXP3} & 3.18 \newline (5.05) & 3.03 \newline (10.08) & 65.46 \newline (3212.16) & 91.81 \newline (3671.66) & 2.62 \newline (2.35) & 7.20 \newline (303.29) & 67.54 \newline (2346.82) & 2.97 \newline (3.99) & 3.06 \newline (2.71) & 77.01 \newline (3135.17) & \textbf{32.39 \newline (1269.33)}\\ 
\hline
\end{tabularx}

  \end{tiny}
  \caption{\capstyle{Tracking error on several video sequences using different
      methods for gaze selection.  The table shows mean tracking error
      as well as the error variance (in brackets) over a single test
      sequence.}}
  \label{tab:experiment2}
\end{table}

%% file: conclusion.tex
\section{Conclusions and Future Work}

We have proposed a decision-theoretic probabilistic graphical model
for joint classification, tracking and planning. The experiments
demonstrate the significant potential of this approach.  We examined
several different strategies for gaze control in both the full and
partial information settings.  We saw that a straightforward
generalization of the full information policy to partial information
gave poor performance and we proposed an alternative method which is
able not only to perform well in the presence of partial information
but also allows us to expand the set of possible fixation points to be
a continuous domain.

There are many routes for further exploration. In this work we
pre-trained the (factored)-RBMs. However, existing particle filtering
and stochastic optimization algorithms could be used to train the RBMs
online. Following the same methodology, we should also be able to
adapt and improve the target templates and proposal distributions over
time. This is essential to extend the results to long video sequences
where the object undergoes significant transformations (e.g.\ as is
done in the predator tracking system~\citep{kalal2010face}).

Deployment to more complex video sequences will require more careful
and thoughtful design of the proposal distributions, transition
distributions, control algorithms, template models, data-association
and motion analysis modules. Fortunately, many of the solutions to
these problems have already been engineered in the computer vision,
tracking and online learning communities. Admittedly, much work
remains to be done.

Saliency maps are ubiquitous in visual attention studies. Here, we
simply used standard saliency tools and motion flow in the
construction of the proposal distributions for particle
filtering. There might be better ways to exploit the saliency maps, as
neurophysiological experiments seem to suggest~\citep{Gottlieb1998}.

One of the most interesting avenues for future work is the
construction of more abstract attentional strategies. In this work, we
focused on attending to regions of the visual field, but clearly one
could attend to subsets of receptive fields or objects in the deep
appearance model.

The current model has no ability to recover from a tracking failure.
It may be possible to use information from the identity pathway (i.e.\
the classifier output) to detect and recover from tracking failure.

A closer examination of the exploration/exploitation tradeoff in the
tracking setting is in order.  For instance, the methods we considered
assume that future rewards are independent of past actions.  This
assumption is clearly not true in our setting, since choosing a long
sequence of very poor fixation points can lead to tracking failure.
We can potentially solve this problem by incorporating the current
tracking confidence into the gaze selection strategy.  This would
allow the exploration/exploitation trade off to be explicitly
modulated by the needs of the tracker, e.g.\ after choosing a poor
fixation point the selection policy could be adjusted temporarily to
place extra emphasis on exploiting good fixation points until
confidence in the target location has been recovered.  Contextual
bandits provide a framework for integrating and reasoning about this
type of side-information in a principled manner.


%% file: ncRBMtrack.bbl
\begin{thebibliography}{41}
\providecommand{\natexlab}[1]{#1}
\providecommand{\url}[1]{\texttt{#1}}
\expandafter\ifx\csname urlstyle\endcsname\relax
  \providecommand{\doi}[1]{doi: #1}\else
  \providecommand{\doi}{doi: \begingroup \urlstyle{rm}\Url}\fi

\bibitem[Auer et~al.(1998{\natexlab{a}})Auer, Cesa-Bianchi, Freund, and
  Schapire]{auer1998gambling}
Auer, P., Cesa-Bianchi, N., Freund, Y., and Schapire, R.E.
\newblock {Gambling in a rigged casino: The adversarial multi-armed bandit
  problem}.
\newblock In \emph{focs}, pp.\  322. Published by the IEEE Computer Society,
  1998{\natexlab{a}}.

\bibitem[Auer et~al.(2001)Auer, Cesa-Bianchi, Freund, and
  Schapire]{auer2001nonstochastic}
Auer, P., Cesa-Bianchi, N., Freund, Y., and Schapire, R.E.
\newblock {The nonstochastic multiarmed bandit problem}.
\newblock \emph{SIAM Journal on Computing}, 32\penalty0 (1):\penalty0 48--77,
  2001.
\newblock ISSN 0097-5397.

\bibitem[Auer et~al.(1998{\natexlab{b}})Auer, Cesa-Bianchi, Freund, and
  Schapire]{Auer:1998}
Auer, Peter, Cesa-Bianchi, Nicol{\`o}, Freund, Yoav, and Schapire, Robert~E.
\newblock Gambling in a rigged casino: the adversarial multi-armed bandit
  problem.
\newblock Technical Report NC2-TR-1998-025, 1998{\natexlab{b}}.

\bibitem[Bazzani et~al.(2010)Bazzani, de~Freitas, and
  Ting]{bazzani2010learning}
Bazzani, L., de~Freitas, N., and Ting, J.A.
\newblock {Learning attentional mechanisms for simultaneous object tracking and
  recognition with deep networks}.
\newblock \emph{NIPS 2010 Deep Learning and Unsupervised Feature Learning
  Workshop}, 2010.

\bibitem[Brochu et~al.(2010)Brochu, Cora, and de~Freitas]{brochu2010tutorial}
Brochu, E., Cora, V.M., and de~Freitas, N.
\newblock {A tutorial on Bayesian optimization of expensive cost functions,
  with application to active user modeling and hierarchical reinforcement
  learning}.
\newblock Technical report, University of British Columbia, 2010.

\bibitem[Colombo(2001)]{Colombo:2001}
Colombo, John.
\newblock The development of visual attention in infancy.
\newblock \emph{Annual Review of Psychology}, pp.\  337--367, 2001.

\bibitem[Doucet et~al.(2001)Doucet, {de Freitas}, and Gordon]{315}
Doucet, A, {de Freitas}, N, and Gordon, N.
\newblock Introduction to sequential {Monte Carlo} methods.
\newblock In Doucet, A, de~Freitas, N, and Gordon, N~J (eds.), \emph{Sequential
  {Monte Carlo} Methods in Practice}. Springer-Verlag, 2001.

\bibitem[Freund \& Schapire(1997)Freund and Schapire]{Freund:1997}
Freund, Yoav and Schapire, Robert~E.
\newblock A decision-theoretic generalization of on-line learning and an
  application to boosting.
\newblock \emph{Journal of Computer and System Sciences}, 55:\penalty0
  119--139, 1997.

\bibitem[Gaborski et~al.(2004)Gaborski, Vaingankar, Chaoji, Teredesai, and
  Tentler]{gaborski2004detection}
Gaborski, R., Vaingankar, V., Chaoji, V., Teredesai, A., and Tentler, A.
\newblock Detection of inconsistent regions in video streams.
\newblock In \emph{Proc. SPIE Human Vision and Electronic Imaging}. Citeseer,
  2004.

\bibitem[Gao et~al.(2007)Gao, Mahadevan, and Vasconcelos]{gao2007discriminant}
Gao, D., Mahadevan, V., and Vasconcelos, N.
\newblock The discriminant center-surround hypothesis for bottom-up saliency.
\newblock \emph{Advances in neural information processing systems}, 20, 2007.

\bibitem[Girard \& Berthoz(2005)Girard and Berthoz]{Girard2005}
Girard, B. and Berthoz, A.
\newblock From brainstem to cortex: Computational models of saccade generation
  circuitry.
\newblock \emph{Progress in Neurobiology}, 77\penalty0 (4):\penalty0 215 --
  251, 2005.

\bibitem[Gottlieb et~al.(1998)Gottlieb, Kusunoki, and Goldberg]{Gottlieb1998}
Gottlieb, Jacqueline~P., Kusunoki, Makoto, and Goldberg, Michael~E.
\newblock The representation of visual salience in monkey parietal cortex.
\newblock \emph{Nature}, 391:\penalty0 481--484, 1998.

\bibitem[Hinton \& Salakhutdinov(2006)Hinton and Salakhutdinov]{hinton2006rdd}
Hinton, GE and Salakhutdinov, RR.
\newblock {Reducing the dimensionality of data with neural networks}.
\newblock \emph{Science}, 313\penalty0 (5786):\penalty0 504--507, 2006.

\bibitem[Isard \& Blake(1996)Isard and Blake]{93}
Isard, M and Blake, A.
\newblock Contour tracking by stochastic propagation of conditional density.
\newblock In \emph{European Computer Vision Conference}, pp.\  343--356, 1996.

\bibitem[Itti et~al.(1998)Itti, Koch, and Niebur]{Itti1998}
Itti, L., Koch, C., and Niebur, E.
\newblock A model of saliency-based visual attention for rapid scene analysis.
\newblock \emph{IEEE Transactions on Pattern Analysis and Machine
  Intelligence}, 20\penalty0 (11):\penalty0 1254 --1259, 1998.

\bibitem[Jones et~al.(1993)Jones, Perttunen, and
  Stuckman]{jones1993lipschitzian}
Jones, D.R., Perttunen, C.D., and Stuckman, B.E.
\newblock {Lipschitzian optimization without the Lipschitz constant}.
\newblock \emph{Journal of Optimization Theory and Applications}, 79\penalty0
  (1):\penalty0 157--181, 1993.
\newblock ISSN 0022-3239.

\bibitem[Kalal et~al.(2010)Kalal, Mikolajczyk, and Matas]{kalal2010face}
Kalal, Z., Mikolajczyk, K., and Matas, J.
\newblock Face-tld: Tracking-learning-detection applied to faces.
\newblock In \emph{Image Processing (ICIP), 2010 17th IEEE International
  Conference on}, pp.\  3789--3792. IEEE, 2010.

\bibitem[Kavukcuoglu et~al.(2009)Kavukcuoglu, Ranzato, Fergus, and
  Le-Cun]{Kavukcuoglu2009}
Kavukcuoglu, K., Ranzato, M.A., Fergus, R., and Le-Cun, Yann.
\newblock Learning invariant features through topographic filter maps.
\newblock In \emph{Computer Vision and Pattern Recognition}, pp.\  1605--1612,
  2009.

\bibitem[Kim et~al.(2008)Kim, Kumar, Pavlovic, and Rowley]{kim2008face}
Kim, M., Kumar, S., Pavlovic, V., and Rowley, H.
\newblock {Face tracking and recognition with visual constraints in real-world
  videos}.
\newblock \emph{IEEE Conf. Computer Vision and Pattern Recognition}, 2008.

\bibitem[Koch \& Ullman(1985)Koch and Ullman]{koch1985shifts}
Koch, C. and Ullman, S.
\newblock Shifts in selective visual attention: towards the underlying neural
  circuitry.
\newblock \emph{Hum Neurobiol}, 4\penalty0 (4):\penalty0 219--27, 1985.

\bibitem[K\"{o}ster \& Hyv\"{a}rinen(2007)K\"{o}ster and
  Hyv\"{a}rinen]{Koster2007}
K\"{o}ster, Urs and Hyv\"{a}rinen, Aapo.
\newblock A two-layer {ICA}-like model estimated by score matching.
\newblock In \emph{International Conference of Artificial Neural Networks},
  pp.\  798--807, 2007.

\bibitem[Larochelle \& Hinton(2010)Larochelle and Hinton]{Larochelle2010}
Larochelle, Hugo and Hinton, Geoffrey.
\newblock Learning to combine foveal glimpses with a third-order {Boltzmann}
  machine.
\newblock In \emph{Neural Information Processing Systems}, 2010.

\bibitem[Lee et~al.(2009)Lee, Grosse, Ranganath, and Ng]{lee2009convolutional}
Lee, H., Grosse, R., Ranganath, R., and Ng, A.Y.
\newblock {Convolutional deep belief networks for scalable unsupervised
  learning of hierarchical representations}.
\newblock In \emph{International Conference on Machine Learning}, 2009.

\bibitem[McNaughton et~al.(2006)McNaughton, Battaglia, Jensen, Moser, and
  Moser]{mcnaughton2006}
McNaughton, Bruce~L., Battaglia, Francesco~P., Jensen, Ole, Moser, Edvard~I.,
  and Moser, May-Britt.
\newblock Path integration and the neural basis of the 'cognitive map'.
\newblock \emph{Nature Reviews Neuroscience}, 7\penalty0 (8):\penalty0
  663--678, 2006.

\bibitem[Okuma et~al.(2004)Okuma, Taleghani, de~Freitas, and Lowe]{Okuma:2004}
Okuma, Kenji, Taleghani, Ali, de~Freitas, Nando, and Lowe, David~G.
\newblock A boosted particle filter: Multitarget detection and tracking.
\newblock In \emph{ECCV}, 2004.

\bibitem[{Olshausen} \& {Field}(1996){Olshausen} and {Field}]{Olshausen1996}
{Olshausen}, B.~A. and {Field}, D.~J.
\newblock {Emergence of simple-cell receptive field properties by learning a
  sparse code for natural images}.
\newblock \emph{Nature}, 381:\penalty0 607--609, 1996.

\bibitem[Olshausen et~al.(1993{\natexlab{a}})Olshausen, Anderson, and
  Van~Essen]{olshausen1993neurobiological}
Olshausen, B.A., Anderson, C.H., and Van~Essen, D.C.
\newblock {A neurobiological model of visual attention and invariant pattern
  recognition based on dynamic routing of information}.
\newblock \emph{The Journal of Neuroscience}, 13\penalty0 (11):\penalty0 4700,
  1993{\natexlab{a}}.
\newblock ISSN 0270-6474.

\bibitem[Olshausen et~al.(1993{\natexlab{b}})Olshausen, Anderson, and
  Essen]{Olshausen1993}
Olshausen, Bruno~A., Anderson, Charles~H., and Essen, David C.~Van.
\newblock A neurobiological model of visual attention and invariant pattern
  recognition based on dynamic routing of information.
\newblock \emph{Journal of Neuroscience}, 13:\penalty0 4700--4719,
  1993{\natexlab{b}}.

\bibitem[Postma et~al.(1997)Postma, van~den Herik, and Hudson]{Postma1997}
Postma, Eric~O., van~den Herik, H.~Jaap, and Hudson, Patrick T.~W.
\newblock {SCAN}: A scalable model of attentional selection.
\newblock \emph{Neural Networks}, 10\penalty0 (6):\penalty0 993 -- 1015, 1997.

\bibitem[Ranzato \& Hinton(2010)Ranzato and Hinton]{Ranzato2010}
Ranzato, M.A. and Hinton, G.E.
\newblock Modeling pixel means and covariances using factorized third-order
  {Boltzmann} machines.
\newblock In \emph{Computer Vision and Pattern Recognition}, pp.\  2551--2558,
  2010.

\bibitem[Rasmussen \& Williams(2006)Rasmussen and
  Williams]{rasmussen2006gaussian}
Rasmussen, C.E. and Williams, C.K.I.
\newblock \emph{{Gaussian processes for machine learning}}.
\newblock Adaptive computation and machine learning. MIT Press, 2006.
\newblock ISBN 9780262182539.
\newblock URL \url{http://books.google.ca/books?id=vWtwQgAACAAJ}.

\bibitem[Rensink(2000)]{Rensink:2000}
Rensink, Ronald~A.
\newblock The dynamic representation of scenes.
\newblock \emph{Visual Cognition}, pp.\  17--42, 2000.

\bibitem[Rosa(2002)]{Rosa2002}
Rosa, M.G.P.
\newblock Visual maps in the adult primate cerebral cortex: Some implications
  for brain development and evolution.
\newblock \emph{Brazilian Journal of Medical and Biological Research},
  35:\penalty0 1485 -- 1498, 2002.

\bibitem[Srinivas et~al.(2010)Srinivas, Krause, Kakade, and
  Seeger]{srinivas2009gaussian}
Srinivas, N., Krause, A., Kakade, S.M., and Seeger, M.
\newblock {Gaussian process optimization in the bandit setting: No regret and
  experimental design}.
\newblock \emph{International Conference on Machine Learning}, 2010.

\bibitem[Swersky et~al.(2010)Swersky, Chen, Marlin, and
  de~Freitas]{Swersky2010}
Swersky, K., Chen, Bo, Marlin, B., and de~Freitas, N.
\newblock A tutorial on stochastic approximation algorithms for training
  restricted {Boltzmann} machines and deep belief nets.
\newblock In \emph{ITA Workshop}, pp.\  1--10, 2010.

\bibitem[Swersky et~al.(2011)Swersky, Buchman, Marlin, and
  de~Freitas]{swersky2011autoencoders}
Swersky, K., Buchman, D., Marlin, B.M., and de~Freitas, N.
\newblock On autoencoders and score matching for energy based models.
\newblock \emph{International Conference in Machine Learning}, 2011.

\bibitem[Taylor et~al.(2010)Taylor, Sigal, Fleet, and Hinton]{Taylor2010}
Taylor, G.W., Sigal, L., Fleet, D.J., and Hinton, G.E.
\newblock Dynamical binary latent variable models for {3D} human pose tracking.
\newblock In \emph{Computer Vision and Pattern Recognition}, pp.\  631--638,
  2010.

\bibitem[Torralba et~al.(2006)Torralba, Oliva, Castelhano, and
  Henderson]{torralba2006contextual}
Torralba, A., Oliva, A., Castelhano, M.S., and Henderson, J.M.
\newblock Contextual guidance of eye movements and attention in real-world
  scenes: The role of global features in object search.
\newblock \emph{Psychological review}, 113\penalty0 (4):\penalty0 766, 2006.

\bibitem[Vincent et~al.(2008)Vincent, Larochelle, Bengio, and
  Manzagol]{vincent2008eac}
Vincent, P., Larochelle, H., Bengio, Y., and Manzagol, P.A.
\newblock {Extracting and composing robust features with denoising
  autoencoders}.
\newblock In \emph{International Conference on Machine Learning}, pp.\
  1096--1103, 2008.

\bibitem[Welling et~al.(2005)Welling, Rosen-Zvi, and Hinton]{welling2005efh}
Welling, M., Rosen-Zvi, M., and Hinton, G.
\newblock {Exponential family harmoniums with an application to information
  retrieval}.
\newblock \emph{Neural Information Processing Systems}, 17:\penalty0
  1481--1488, 2005.

\bibitem[Zhang et~al.(2009)Zhang, Tong, and Cottrell]{zhang2009sunday}
Zhang, L., Tong, M.H., and Cottrell, G.W.
\newblock Sunday: Saliency using natural statistics for dynamic analysis of
  scenes.
\newblock In \emph{Proceedings of the 31st Annual Cognitive Science Conference,
  Amsterdam, Netherlands}. Citeseer, 2009.

\end{thebibliography}
